\documentclass[lettersize,journal]{IEEEtran}
\usepackage{amsmath,amsfonts}
\usepackage{algorithmic}
\usepackage{algorithm}
\usepackage{array}
\usepackage[caption=false,font=normalsize,labelfont=sf,textfont=sf]{subfig}
\usepackage{textcomp}
\usepackage{stfloats}
\usepackage{url}
\usepackage{verbatim}
\usepackage{graphicx}
\usepackage{cite}

\usepackage {makecell}
\usepackage{multirow}
\usepackage{amsmath,amsfonts}
\usepackage{algorithmic}
\usepackage{algorithm}
\usepackage{array}
\usepackage{xurl}
\usepackage{dsfont}
\usepackage{ulem}
\usepackage{amsfonts}
\usepackage{mathrsfs}
\usepackage{amssymb}
\usepackage{multirow}
\usepackage{colortbl}
\usepackage{booktabs}
\usepackage{scalerel}
\usepackage{tikz}
\usetikzlibrary{svg.path}
\definecolor{orcidlogocol}{HTML}{A6CE39}
\tikzset{
	orcidlogo/.pic={
		\fill[orcidlogocol] svg{M256,128c0,70.7-57.3,128-128,128C57.3,256,0,198.7,0,128C0,57.3,57.3,0,128,0C198.7,0,256,57.3,256,128z};
		\fill[white] svg{M86.3,186.2H70.9V79.1h15.4v48.4V186.2z}
		svg{M108.9,79.1h41.6c39.6,0,57,28.3,57,53.6c0,27.5-21.5,53.6-56.8,53.6h-41.8V79.1z M124.3,172.4h24.5c34.9,0,42.9-26.5,42.9-39.7c0-21.5-13.7-39.7-43.7-39.7h-23.7V172.4z}
		svg{M88.7,56.8c0,5.5-4.5,10.1-10.1,10.1c-5.6,0-10.1-4.6-10.1-10.1c0-5.6,4.5-10.1,10.1-10.1C84.2,46.7,88.7,51.3,88.7,56.8z};
	}
}
\newcommand\orcidicon[1]{\href{https://orcid.org/#1}{\mbox{\scalerel*{
				\begin{tikzpicture}[yscale=-1,transform shape]
					\pic{orcidlogo};
				\end{tikzpicture}
			}{|}}}}
\usepackage{hyperref}
\hypersetup{
	colorlinks=true,
	linkcolor=blue,
	filecolor=black,      
	urlcolor=black,
	citecolor=purple
}
\usepackage{tikz-network}

\newcommand{\eg}{\textit{e.g. }}
\newcommand{\etal}{\textit{et al. }}

\newcommand{\settablefont}{\fontsize{8}{15}\selectfont}

\begin{document}
	
	\title{SNE-RoadSegV2: Advancing Heterogeneous Feature Fusion and Fallibility Awareness for Freespace Detection}

	\author{
		Yi Feng$^{\orcidicon{0009-0005-4885-0850}\,}$, 
		Yu Ma$^{\orcidicon{0009-0000-8536-5444}\,}$, 
		Qijun Chen,~\IEEEmembership{Senior Member,~IEEE},\\ \ \ \ Ioannis Pitas,~\IEEEmembership{Life Fellow,~IEEE}, and Rui Fan$^{\orcidicon{0000-0003-2593-6596}\,}$,~\IEEEmembership{Senior Member,~IEEE}
		
		\thanks{Yi Feng, Yu Ma, Qijun Chen, and Rui Fan are with the College of Electronics \& Information Engineering, Shanghai Research Institute for Intelligent Autonomous Systems, the State Key Laboratory of Intelligent Autonomous Systems, and Frontiers Science Center for Intelligent Autonomous Systems, Tongji University, Shanghai 201804, P. R. China (e-mail: \{fengyi0109, mayu2002, qjchen, rfan\}@tongji.edu.cn). }
		\thanks{Ioannis Pitas is with the Department of Informatics, University of Thessaloniki, 541 24 Thessaloniki, Greece (e-mail: pitas@csd.auth.gr).}
		
	}

	
	\maketitle
	
	\begin{abstract}
		Feature-fusion networks with duplex encoders have proven to be an effective technique to solve the freespace detection problem. However, despite the compelling results achieved by previous research efforts, the exploration of adequate and discriminative heterogeneous feature fusion, as well as the development of fallibility-aware loss functions remains relatively scarce. This paper makes several significant contributions to address these limitations: (1) It presents a novel heterogeneous feature fusion block, comprising a holistic attention module, a heterogeneous feature contrast descriptor, and an affinity-weighted feature recalibrator, enabling a more in-depth exploitation of the inherent characteristics of the extracted features, (2) it incorporates both inter-scale and intra-scale skip connections into the decoder architecture while eliminating redundant ones, leading to both improved accuracy and computational efficiency, and (3) it introduces two fallibility-aware loss functions that separately focus on semantic-transition and depth-inconsistent regions, collectively contributing to greater supervision during model training. Our proposed heterogeneous feature fusion network (SNE-RoadSegV2), which incorporates all these innovative components, demonstrates superior performance in comparison to all other freespace detection algorithms across multiple public datasets. Notably, it ranks the 1st on the official KITTI Road benchmark. 
	\end{abstract}
	
	\begin{IEEEkeywords}
		Freespace detection, Feature fusion, semantic segmentation, deep learning.
	\end{IEEEkeywords}
	
	\section{Introduction}
	\label{Sect.intro}
	\IEEEPARstart{A}{s} a vital piece of the autonomous driving puzzle, reliable collision-free space (freespace for short) detection holds significant importance in autonomous driving systems, as it directly impacts a vehicle's ability to make informed decisions and ensure dependable navigation \cite{fan2020sne}. In recent years, freespace detection has attracted considerable attention in research, with ongoing efforts aimed at addressing corner cases within complex and dynamic environments. Nevertheless, regardless of whether the approach is explicit programming-based or data-driven, the utilization of 3D information is growing in significance for freespace detection, primarily due to the valuable spatial geometry information it provides \cite{chen2019progressive}.
	Feature-fusion networks with duplex encoders, designed to extract heterogeneous features from multiple data sources or modalities and fuse them to provide a more comprehensive understanding of the environment, have emerged as a viable solution to tackle this problem \cite{hazirbas2017fusenet, ha2017mfnet}.
	
	The performance of a feature-fusion freespace detection network depends not only on the input data type but also on the manner in which these features are fused \cite{zhou2022fanet, zhou2022canet}. 
	A current bottleneck lies in the simplistic and indiscriminate fusion of heterogeneous features, often causing conflicting feature representations and erroneous detection results \cite{chen2019progressive}. Taking the SNE-RoadSeg series \cite{fan2020sne,wang2021sne} as an example, their adopted feature fusion strategy essentially performs an element-wise summation between RGB and surface normal feature maps at each stage, neglecting the inherent differences in feature characteristics and their respective reliability \cite{zou2023dual}. Furthermore, as the network goes deeper, such an asymmetric feature fusion strategy tends to diminish the proportion of RGB features in the decoder's input. This, in turn, leads to unsatisfactory performance, particularly in areas such as pavements or other lanes, where surface normals closely resemble those of freespace, demanding a greater reliance on color or textural information. 
	
	In the decoder aspect, we observe two phenomena: (1) inter-scale skip connections provide an advantage in achieving more comprehensive feature decoding, primarily due to their capability to capture both fine-grained and coarse-grained details \cite{huang2020unet}, but unfortunately they are underutilized; (2) the currently adopted intra-scale skip connections appear to be excessively redundant for this task. Moreover, the utilization of pixel-wise binary cross-entropy (BCE) loss has been a common practice in freespace detection. It is, nevertheless, regrettable that no prior endeavors have been undertaken to delve into the fallible cases, notably the misclassifications occurring near semantic-transition and depth-inconsistent regions.
	
	\begin{figure}[!t]
		\centering
		\includegraphics[width=0.49\textwidth]{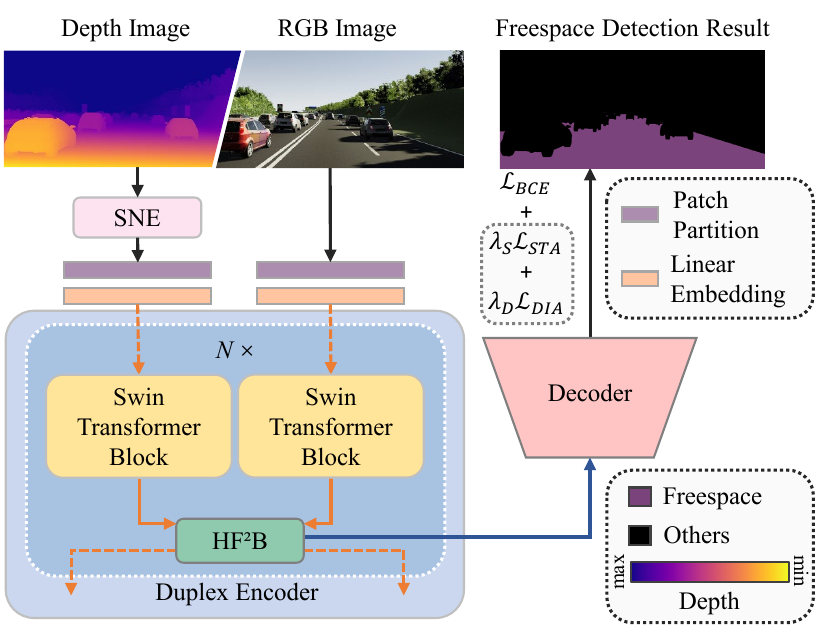}
		\caption{An overview of our proposed SNE-RoadSegV2. }
		\label{fig.overview}
	\end{figure}
	
	To address the aforementioned limitations, we first dive deeper into the discriminative feature fusion strategies presented in recent universal semantic segmentation studies \cite{fan2020sne, wang2021sne}. Subsequently, we introduce a novel heterogeneous feature fusion block (HF$^2$B) to process the RGB and surface normal features, which are encoded using two independent Swin Transformer backbones \cite{liu2021swin}. Our technical contributions in this part are three-fold: (1) a holistic attention module (HAM) to model the interdependencies between heterogeneous features across three dimensions (spatial, channel, and scale); (2) a heterogeneous feature contrast descriptor (HFCD) to effectively underscore both the shared and unique characteristics within the holistically-attentive features; (3) an affinity-weighted feature recalibrator (AWFR) to jointly emphasize and suppress heterogeneous features prior to their input into the decoder. Additionally, we contribute to a lightweight yet more effective decoder, which incorporates inter-scale skip connections while pruning redundant ones. Our decoder demonstrates greater feature decoding capabilities, while simultaneously reducing computational complexity. Finally, we design two new loss functions based on semantic annotations and depth data to provide deeper supervision during our model training process. This contribution also results in improved overall performance, particularly in error-prone areas. The effectiveness of each contribution is validated through extensive experiments conducted across public datasets. Excitingly, our proposed freespace detection framework SNE-RoadSegV2, as shown in Fig. \ref{fig.overview}, which incorporates all these innovative components, wins the \textbf{1st place} on the KITTI Road leaderboard \cite{geiger2012we}. In a nutshell, our contributions are as follows:
	\begin{enumerate}
		\item We propose SNE-RoadSegV2, a novel feature-fusion freespace detection approach, achieving state-of-the-art (SoTA) performance across multiple public datasets. Our source code will be released upon publication. 
		
		\item We introduce HF$^2$B, consisting of an HAM, an HFCD, and an AWFR, for comprehensive heterogeneous feature description, recalibration, and fusion, resulting in more coherent feature representations.  
		
		\item We design a lightweight yet more effective decoder, incorporating inter-scale skip connections while pruning redundant intra-scale ones, demonstrating greater efficacy and computational efficiency.
		
		\item We develop two novel fallibility-aware loss functions, which focus particularly on reducing misclassifications in semantic-transition and depth-inconsistent regions, leading to improved overall performance. 	
	\end{enumerate}
	
	\begin{figure*}[!t]
		\centering
		\includegraphics[width=0.99\textwidth]{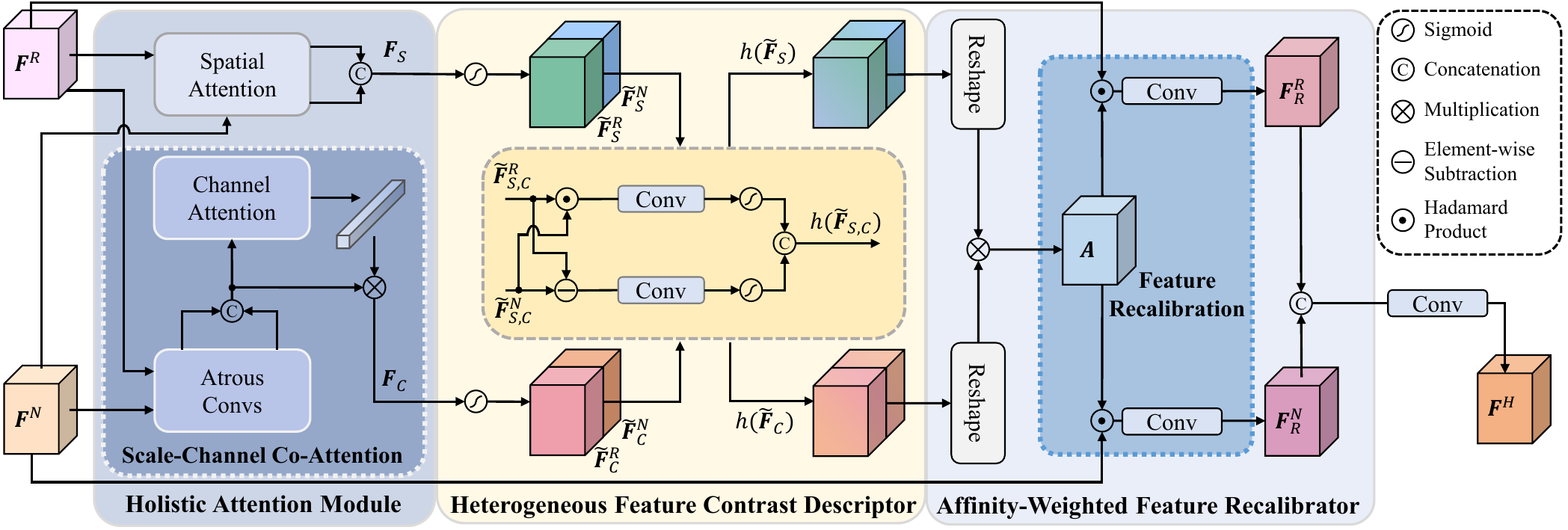}
		\caption{An illustration of our proposed heterogeneous feature fusion block, consisting of (1) a holistic attention module, (2) a heterogeneous feature contrast descriptor, and (3) an affinity-weighted feature recalibrator.}
		\label{fig.fusion}
	\end{figure*}

	\section{Related Work}
	\label{sec:related_work}
	
	\subsection{Data-Driven Freespace Detection}
	
	While it is feasible to employ universal semantic segmentation networks \cite{long2015fully, ronneberger2015u, badrinarayanan2017segnet, chen2017deeplab} for this task, it has been observed that task-specific approaches \cite{li2023roadformer, fan2020sne, wang2021sne} consistently deliver superior performance. Early task-specific freespace detection approaches \cite{alvarez2012road, xiao2016monocular, Brust2015ConvolutionalPN, levi2015stixelnet} thoroughly rely on RGB images and were found to be highly sensitive to environmental factors, notably illumination and weather conditions \cite{fan2020sne}. Given the increased prevalence of range sensors, particularly LiDARs, feature-fusion networks \cite{chen2019progressive, khan2022lrdnet, gu2021cascaded} have emerged as a more robust choice in this domain. In terms of network architecture, these approaches are characterized by duplex-encoder architectures \cite{fan2020sne,li2023roadformer}, where each encoder extracts hierarchical features from a specific data source or modality. The extracted heterogeneous features are subsequently fused, enabling the network to gain a more comprehensive understanding of the environment \cite{chen2019progressive}.  As for the input data, the most commonly used spatial geometric information includes depth/disparity maps \cite{chang2022fast, sun2019reverse}, LiDAR point clouds \cite{chen2019progressive, khan2022lrdnet}, and surface normal information \cite{fan2020sne, wang2021sne, li2023roadformer}. Extensive experiments conducted in previous studies \cite{fan2020sne, wang2021sne, li2023roadformer} have conclusively demonstrated that surface normals provide the most informative spatial geometric information for freespace detection, owing to their representation of planar characteristics. Therefore, in this paper, we adopt the pipeline introduced in \cite{fan2020sne, wang2021sne, li2023roadformer}, which utilizes a duplex-encoder architecture to extract heterogeneous features from RGB images and surface normal information. However, it is important to note that our focus differs from these previous work. Our emphasis lies in designing heterogeneous feature fusion strategies, developing a lightweight yet more effective decoder, and introducing task-specific loss functions.

	\subsection{Heterogeneous Feature Fusion}
	\label{sec.heterogeneous_feature_fusion}
	Heterogeneous feature fusion plays a pivotal role in various computer vision tasks, such as salient object detection  \cite{wei2020f3net, chen2020bi, pang2020hierarchical} and scene parsing  \cite{qiu2021semantic, sun2020real, caltagirone2019lidar}. Although this topic may not have received extensive attention in freespace detection, it is worth noting that there have been notable prior works proposed in the broader field of semantic segmentation. For instance, Wei \etal \cite{wei2020f3net} proposed the cross feature module (CFM), capable of refining features at multiple levels while simultaneously suppressing background noise. Additionally, the separation-and-aggregation gate (SAGate) \cite{chen2020bi} is another general-purpose heterogeneous feature fusion method that incorporates complementary information through feature recalibration and aggregation to generate selective representations for segmentation. Moreover, Pang \etal \cite{chen2020bi} introduced the dynamic dilated pyramid module (DDPM), which generates adaptive kernels for efficient feature decoding. As these prior studies generally overlook the appropriate discrimination between the inherent differences between heterogeneous features, our primary focus in this paper is directed towards addressing this aspect.
	
	\subsection{Attention Mechanisms}
	\label{sec.review_attention}
	Attention mechanisms are vital components within modern deep learning models, allowing for effective concentration on specific elements of input data, ultimately leading to a more comprehensive understanding of the environment \cite{hu2018squeeze, xu2015show, vaswani2017attention}. As a representative example, the squeeze-and-excitation network (SENet) \cite{hu2018squeeze} dynamically recalibrates channel-wise feature responses by explicitly modeling dependencies between channels, enabling the network to emphasize informative channels while suppressing less relevant ones. In addition to channel attention, the convolutional block attention module (CBAM) \cite{woo2018cbam} introduces attention from another dimension -- spatial. This lightweight and highly compatible module sequentially computes channel and spatial attention maps and multiplies them with the input feature maps to achieve more adaptive feature refinement. On the other hand, Swin Transformer \cite{liu2021swin} is a general-purpose Transformer backbone developed specifically for fundamental computer vision tasks. It is highly regarded for its hierarchical representation learning approach, which computes self-attention locally within non-overlapping shifted windows. This innovative design contributes to its compelling performance in applications including image classification, object detection, and semantic segmentation. In this paper, we first extend CBAM to three dimensions: the original two plus scale. Moreover, we use two Swin Transformers as the backbone networks in our duplex encoder, and comprehensive experiments in the supplement provide evidence of its superior performance compared to CNNs.

	\section{Methodology}
	\label{sec:method}
	
	\subsection{Architecture Overview}
	\label{Sect.architecture}
	
	Fig. \ref{fig.overview} provides readers with an overview of the SNE-RoadSegV2 architecture, consisting of three key elements: duplex feature embedding, heterogeneous feature fusion, and lightweight yet effective feature decoding. A pair of input RGB image $\boldsymbol{I}^{R}$ and surface normal map $\boldsymbol{I}^N$, translated from a depth image $\boldsymbol{I}^D$ using a surface normal estimator (SNE) \cite{fan2020sne}, are first tokenized into non-overlapping patches and transformed into a high-dimensional feature space through a trainable linear projection in the patch embedding module \cite{liu2021swin}. The embedded features are subsequently fed into the Swin Transformer blocks \cite{liu2021swin} to produce hierarchical heterogeneous features $\mathcal{F}^{R} =\{\boldsymbol{F}^{R}_{1},..., \boldsymbol{F}^{R}_{k}\}$ and $\mathcal{F}^{N} =\{\boldsymbol{F}^{N}_{1},..., \boldsymbol{F}^{N}_{k}\}$. Each pair of heterogeneous features $\boldsymbol{F}^{R,N}_{i}\in \mathbb{R}^{C\times H\times W}$ undergoes a comprehensive fusion process through HF$^2$B. Finally, a lightweight yet more effective decoder which incorporates both inter-scale and intra-scale skip connections is designed to further boost the efficiency and accuracy of freespace detection. The proposed architecture is trained by minimizing a loss with fallibility-awareness incorporated. The following subsections will provide a detailed explanation of the HF$^2$B, decoder, and loss functions in sequence.
	
	\subsection{Heterogeneous Feature Fusion Block}
	\label{Sect.heterogeneous_feature_fusion}
	The core problem of feature encoding for freespace detection revolves around the effective fusion of heterogeneous features extracted from various data sources. As detailed in Sec. \ref{Sect.intro}, prior investigations, \eg, \cite{fan2020sne, wang2021sne, chang2022fast, khan2022lrdnet} fused these heterogeneous features without appropriate discrimination between their inherent differences. Our HF$^2$B is, thus, designed to overcome this limitation. As depicted in Fig. \ref{fig.fusion}, meaningful heterogeneous features are first selectively emphasized and suppressed across three dimensions (spatial, channel, and scale) through an HAM. These features are further enhanced through an HFCD, which improves the representations of their shared and distinct characteristics. The original heterogeneous features are ultimately weighted through an AWFR to emphasize the aspects that are important to both or either of the features.
	
	\subsubsection{Holistic Attention Module}
	\label{Sect.holistic_attention_module}
	Before contrasting and fusing heterogeneous features, it is imperative to emphasize or attenuate specific spatial regions and channels across multiple scales \cite{woo2018cbam, chen2017deeplab}. Drawing inspiration from the CBAM \cite{woo2018cbam}, we first apply spatial attention to both RGB and surface normal feature maps $\boldsymbol{F}^{R}$ and $\boldsymbol{F}^{N}$, resulting in spatially weighted feature maps $\boldsymbol{F}_S^{R}$ and $\boldsymbol{F}_S^{N}$, respectively, which are then concatenated to form $\boldsymbol{F}_S$:
	\begin{equation}
		\boldsymbol{F}_{S}
		\triangleq
		\left[
		\boldsymbol{F}_S^{R};
		\boldsymbol{F}_S^{N}
		\right]
		= 
		\left[
		{f}_{s}(\boldsymbol{F}^{R});
		{f}_{s}(\boldsymbol{F}^{N})
		\right]
		\in \mathbb{R}^{2C\times H \times W},
		\label{eq.fs}
	\end{equation}
	where ${f}_{s}(\cdot)$ denotes the spatial attention operation, allowing the model to prioritize important regions of an image while suppressing less relevant areas. 
	
	As illustrated in Fig. \ref{fig.fusion}, another branch of HAM incorporates multi-scale contextual information along with channel attention. A series of atrous convolutional layers \cite{chen2017deeplab} are initially employed to generate features $\boldsymbol{F}_A$ with progressively expanded receptive fields:
	\begin{equation}
		\boldsymbol{F}_A
		=
		\left[
		{f}_a(\boldsymbol{F}^R); {f}_a(\boldsymbol{F}^N)
		\right]
		\in \mathbb{R}^{2C\times H \times W},
		\label{eq.fa}
	\end{equation}
	where $f_a(\cdot)$ denotes the multi-scale context aggregation operation. A channel attention operation $f_c$ is subsequently applied to further model the interdependencies between heterogeneous features at both scale and channel levels, resulting in scale-channel attentive feature maps $\boldsymbol{F}_C$ as follows:
	\begin{equation}
		\boldsymbol{F}_C
		\triangleq
		\left[
		\boldsymbol{F}_{C}^{R};
		\boldsymbol{F}_{C}^{N}
		\right]
		=
		{f}_c
		\left(
		\boldsymbol{F}_A
		\right)
		\in \mathbb{R}^{2C\times H \times W}.
		\label{eq.fc}
	\end{equation}
	The expressions of $f_s(\cdot)$, $f_a(\cdot)$, and $f_c(\cdot)$ are detailed in the supplement. 
	In contrast to prior arts \cite{woo2018cbam,hu2018squeeze} which exclusively focus on channel attention and overlook the consideration of multiple scales, our designed scale-channel co-attention mechanism provides a more comprehensive perspective on heterogeneous features. 
	
	\subsubsection{Heterogeneous Feature Contrast Descriptor}
	\label{Sect.feature_contrast_module}
	
	After modeling the interdependencies of the heterogeneous features across three separate dimensions using (\ref{eq.fs})-(\ref{eq.fc}), we further explore the way to contrast these features in a more comprehensive and effective fashion. Unlike relevant prior arts, \eg, CFM \cite{wei2020f3net}, SAGate \cite{chen2020bi}, and DDPM \cite{pang2020hierarchical}, which primarily emphasize feature commonality, our investigation delves deeper into both their shared and distinct characteristics. To this end, we first normalize the spatial-attentive and scale-channel co-attentive features $\boldsymbol{F}_{S}$ and $\boldsymbol{F}_{C}$ using a sigmoid function $\sigma(\cdot)$, yielding $\tilde{\boldsymbol{F}}_{S}=[\tilde{\boldsymbol{F}}_S^{R};\tilde{\boldsymbol{F}}_S^{N}]$ and $\tilde{\boldsymbol{F}}_{C}=[\tilde{\boldsymbol{F}}_C^{R};\tilde{\boldsymbol{F}}_C^{N}]$, respectively. Performing an element-wise product operation between $\tilde{\boldsymbol{F}}_{S,C}^{R}$ and $\tilde{\boldsymbol{F}}_{S,C}^{N}$ activates jointly-emphasized parts between heterogeneous features, while performing an element-wise subtraction operation between $\tilde{\boldsymbol{F}}_{S,C}^{R}$ and $\tilde{\boldsymbol{F}}_{S,C}^{N}$ activates features that are important only in either the RGB image or the surface normal map. Upon such inspiration, we formulate our HFCD as follows:
	\begin{equation}
		\begin{split}
			{h}
			\big(
			\tilde{\boldsymbol{F}}_{S,C}
			\big) 
			=
			\big[
			w_{h,1}
			\big(
			\tilde{\boldsymbol{F}}_{S,C}^{R}\odot \tilde{\boldsymbol{F}}_{S,C}^{N}
			\big); 
			w_{h,2}
			\big(
			\tilde{\boldsymbol{F}}_{S,C}^{R}\ominus \tilde{\boldsymbol{F}}_{S,C}^{N}
			\big)
			\big],
		\end{split}
	\end{equation}
	where $\odot$ refers to the Hadamard product, $\ominus$ denotes element-wise subtraction, and $w_{h,i}$ (with its expression provided in the supplement) represents a combination of convolutional, batchnorm, and sigmoid layers. As shown in Fig. \ref{fig.fusion}, the described heterogeneous feature contrast ${h}\big(\tilde{\boldsymbol{F}}_{S,C}\big)$ is subsequently utilized to construct an affinity volume $\boldsymbol{A}$ for further feature recalibration. 
	
	\subsubsection{Affinity-Weighted Feature Recalibrator}
	\label{Sect.affinity_module}
	Constructing a volume that contains element-wise weights to jointly recalibrate (emphasize and de-emphasize) heterogeneous features is another significant contribution in our designed HF$^2$B. As $h(\tilde{\boldsymbol{F}}_{S})$ and $h(\tilde{\boldsymbol{F}}_{C})$ describe the contrasting aspects between the heterogeneous features at the spatial and scale-channel levels, respectively, we employ these two volumes to construct an affinity volume $\boldsymbol{A}\in \mathbb{R}^{C\times H\times W}$ as follows:
	\begin{equation}
		\boldsymbol{A} = 
		\operatorname{Reshape}
		\Big(
		{h}
		(
		\tilde{\boldsymbol{F}}_S
		)
		\Big) 
		\operatorname{Reshape}
		\Big(
		{h}
		(
		\tilde{\boldsymbol{F}}_C
		)
		\Big)
		^\top,
		\label{eq.aff}
	\end{equation}
	which provides the original heterogeneous features $\boldsymbol{F}^R$ and $\boldsymbol{F}^N$ with element-wise weights between 0 and 1. Specifically, a higher affinity value indicates a greater importance of that element in both types of feature maps, an intermediate affinity value indicates a greater importance of element in either of the feature maps, while a lower affinity value indicates a redundant element in both types of feature maps that should be neglected. Finally, $\boldsymbol{F}^R$ and $\boldsymbol{F}^N$ are weighted by $\boldsymbol{A}$ to form $\boldsymbol{F}^R_R$ and $\boldsymbol{F}^N_R$ for the next stage of feature encoding, and then concatenated to generate the recalibrated heterogeneous features as follows:
	\begin{equation}
		\begin{aligned}
			\boldsymbol{F}^H=
			w_a
			\Big(
			\big[
			\underbrace{w_{r}\big(\boldsymbol{F}^{R}\odot\boldsymbol{A}\big)}_{\boldsymbol{F}^{R}_{R}}
			;
			\underbrace{w_{n}\big(\boldsymbol{F}^{N}\odot\boldsymbol{A}\big)}_{\boldsymbol{F}^{N}_{R}}
			\big]
			\Big)
			\in \mathbb{R}^{C\times H\times W},
		\end{aligned}
	\end{equation}
	where $w_r$, $w_n$, and $w_a$ denote convolutional layers. Compared to other SoTA heterogeneous feature fusion strategies, our proposed HF$^2$B adaptively assigns weights to the original features, taking into account both the elements of agreement and disagreement in importance as determined by HFCD and AWFR. Such a way of seeking common ground and while preserving differences enhances the comprehensiveness of feature fusion for freespace detection. The superior performance of HF$^2$B and the effectiveness of its each component are demonstrated in Sec. \ref{Sect.ablation}.
	
	\subsection{Lightweight yet More Effective Decoder}
	\label{Sect.decoder}
\begin{figure*}[!t]
		\centering
		\includegraphics[width=0.90\textwidth]{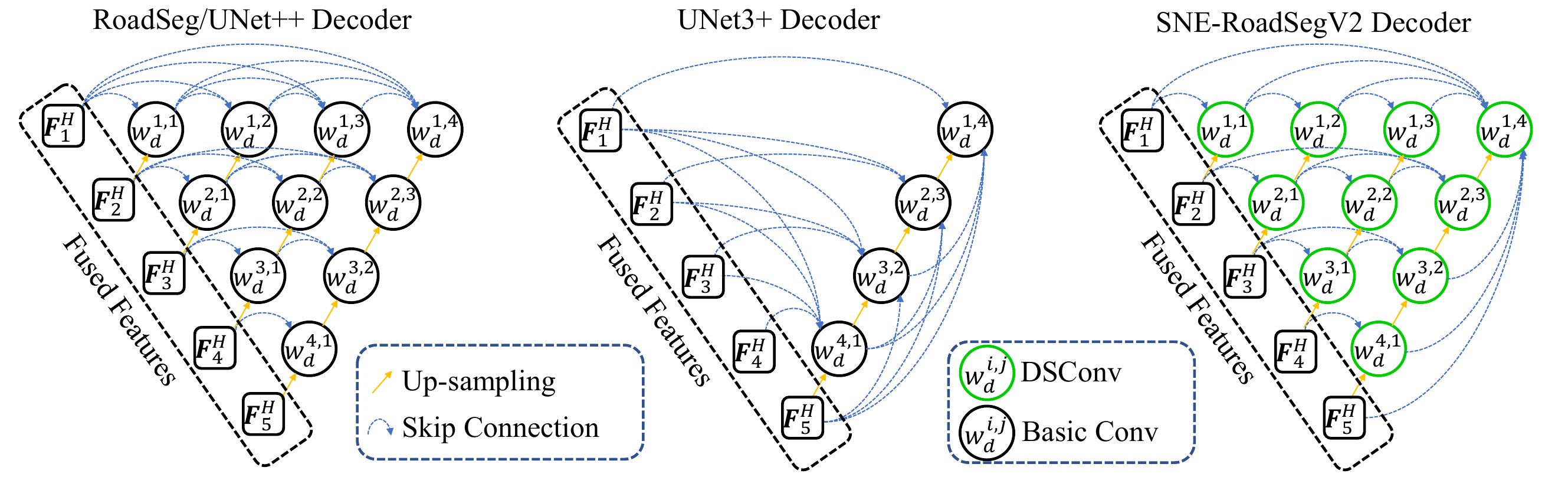}
		\caption{Comparisons among decoder architectures of RoadSeg/UNet++, UNet3+, and our proposed SNE-RoadSegV2. $w_d^{i,j}$ denotes a basic convolutional layer (Basic Conv) or depth-wise separable convolutional layer (DSConv) with batchnorm and sigmoid layers.}
		\label{fig.decoder_compare}
	\end{figure*}

	Fig. \ref{fig.decoder_compare} illustrates the decoder architectures of RoadSeg \cite{fan2020sne} (identical to UNet++ \cite{zhou2019unet++}), UNet3+ \cite{huang2020unet}, and our proposed SNE-RoadSegV2. As claimed in \cite{huang2020unet}, UNet++ fails to effectively utilize multi-scale features, and UNet3+ was designed specifically to resolve this limitation. However, before designing our decoder, we must address the question: does UNet3+ consistently outperform UNet++ in the freespace detection task? Through an extensive series of experiments with both CNN and Transformer backbones, as detailed in the supplement, we regret to report that the answer is negative. It is our contention that both the intra-scale skip connections in UNet++ and the inter-scale skip connections in UNet3+ remain indispensable. Thus, we design the SNE-RoadSegV2 decoder, which combines the strengths of both UNet++ and UNet3+, through enormous experimental efforts. As shown in Fig. \ref{fig.decoder_compare}, we maintain skip connections only from a given node to its adjacent and final nodes at each stage. This modification reduces redundant information propagation without compromising decoder performance. Additionally, we adopt the inter-scale skip connections used in UNet3+ to capture both fine-grained and coarse-grained details. Moreover, we replace the basic convolutions in the decoder with depth-wise separable convolutions \cite{chollet2017xception} to further reduce its computational complexity. Sec. \ref{Sect.ablation} quantitatively demonstrates that our SNE-RoadSegV2 decoder outperforms UNet++ and UNet3+ in terms of both efficiency and accuracy. Despite its superior performance, we consider the decoder as an experimental contribution.
	
	\subsection{Fallibility-Aware Loss Functions}
	\label{Sect.loss_functions}
	
	The pixel-wise BCE loss has been a common choice in previous studies as the primary criterion to supervise model training. Nonetheless, it is unfortunate that these previous efforts have not considered the specific characteristics of real-world driving scenarios. Misclassifications are frequently observed near the transition regions between different semantic categories \cite{wei2020f3net}. In addition, the depth data are not explicitly utilized to supervise model training. Hence, we propose a novel modification to the conventional BCE loss function $\mathcal{L}_{BCE}$ by introducing two weighting factors that prioritize these error-prone regions. 
	The overall loss is formulated as follows:
	\begin{equation}
		\mathcal{L} = \mathcal{L}_{BCE} + \lambda_{S}\mathcal{L}_{STA} + \lambda_{D}\mathcal{L}_{DIA},
		\label{eq.wholeLoss}
	\end{equation}
	where $\lambda_{S}$ and $\lambda_{D}$ balance the semantic transition-aware loss $\mathcal{L}_{STA}$ and the depth inconsistency-aware loss $\mathcal{L}_{DIA}$, respectively. The ablation studies on their individual efficacy and the selection of hyper-parameters $\lambda_{S}$ and $\lambda_{D}$ are provided in Sec. \ref{Sect.ablation}.

	\subsubsection{Semantic Transition-Aware Loss}
	\label{Sect.sta_loss}
	A transition region between different semantic categories can be considered as a mixture of semantic labels. For each pixel $\boldsymbol{q}$ with a neighborhood system $\mathcal{N}_{\boldsymbol{q}}=\mathcal{F}_{\boldsymbol{q}} \cup \mathcal{B}_{\boldsymbol{q}}$, where $\mathcal{F}_{\boldsymbol{q}}$ and $\mathcal{B}_{\boldsymbol{q}}$ denote the foreground (freespace) and background (others) sets, respectively, we determine its likelihood of belonging to a semantic transition region using the following expression:
	\begin{equation}
		\omega_{S}(\boldsymbol{q}) = \cos (\pi \bigg|\frac{\sum_{ {\boldsymbol{p}}} I_{\mathcal{F}_{\boldsymbol{q}}}(\boldsymbol{p}) }{\sum_{{\boldsymbol{p}}} I_{\mathcal{N}_{\boldsymbol{q}}}(\boldsymbol{p})}-\frac{1}{2} \bigg| )\in[0,1],
		\label{eq.omega_s}
	\end{equation}
	where $I(\cdot)$ is the indicator function, and ${\boldsymbol{p}} \in \mathcal{N}_{\boldsymbol{q}}$. (\ref{eq.omega_s}) approaches 0 when either $\mathcal{F}_{\boldsymbol{q}}$ or $\mathcal{B}_{\boldsymbol{q}}$ is close to being an empty set, and approaches 1 vice versa. Substituting (\ref{eq.omega_s}) into the standard BCE loss yields $\mathcal{L}_{STA}$ as follows:
	\begin{equation}
		\mathcal{L}_{STA} = -\sum_{\boldsymbol{q}} \omega_{S}(\boldsymbol{q}) \Big(
		y_{\boldsymbol{q}} \log p_{\boldsymbol{q}}+  (1-y_{\boldsymbol{q}})  \log(1-p_{\boldsymbol{q}} )
		\Big),
		\label{eq.sta_loss}
	\end{equation}
	where $y_{\boldsymbol{q}}\in \{0,1\}$ denotes the ground-truth label of $\boldsymbol{q}$ (1 for freespace, and 0 otherwise), while $p_{\boldsymbol{q}}\in[0,1]$ indicates the probability that $\boldsymbol{q}$ belongs to the freespace category. Our supplement provides details on the selection of $\mathcal{N}_{\boldsymbol{q}}$ radius.
	
	\subsubsection{Depth Inconsistency-Aware Loss}
	\label{Sect.daa_loss}
	When depth images are available, it is also advantageous to leverage these data to improve network training via an adaptive loss function. Surprisingly, prior research efforts have not explored this aspect. 
	
	Let $\tilde{\boldsymbol{Q}}\in\mathbb{R}^{3\times N}$ be a matrix storing the homogeneous coordinates $\tilde{\boldsymbol{q}}$ of the predicted freespace pixels $\boldsymbol{q}$. Considering that the height between the camera and the ground plane remains theoretically constant in each image, when disregarding the camera pitch angle, we aggregate the $y$-coordinates of these pixels to derive a theoretical camera height, denoted as $\hat{y}$, using the following expression:
	\begin{equation}
		\hat{y}=[0,\frac{1}{N},0]\boldsymbol{K}^{-1}\tilde{\boldsymbol{Q}}\boldsymbol{z},
		\label{eq.yhat}
	\end{equation}
	where $\boldsymbol{z}\in\mathbb{R}^{N}$ stores the depth values $\boldsymbol{I}^D(\boldsymbol{q})$. A weight $\omega_{D}(\boldsymbol{q})$ measuring the depth inconsistency can then be yielded as follows:
	\begin{equation}
		\omega_{D}(\boldsymbol{q}) = 1 - \exp\bigg( - \bigg| \frac{\hat{y}}{[0,1,0]\boldsymbol{K}^{-1}\tilde{\boldsymbol{q}}} - \boldsymbol{I}^D(\boldsymbol{q}) \bigg| \bigg) \in[0,1),
		\label{eq.w_D}
	\end{equation}
	which approaches 1 when depth is inconsistent, and approaches 0 vice versa. $\mathcal{L}_{DIA}$ is thus formulated as follows:
	\begin{equation}
		\mathcal{L}_{DIA} = 
		-\sum_{\boldsymbol{q}} \omega_{D}(\boldsymbol{q})
		\Big(
		y_{\boldsymbol{q}} \log p_{\boldsymbol{q}}+  (1-y_{\boldsymbol{q}})  \log(1-p_{\boldsymbol{q}} )
		\Big).
		\label{eq.dia}
	\end{equation}
	
	\begin{figure*}[!t]
		\centering
		\includegraphics[width=0.99\textwidth]{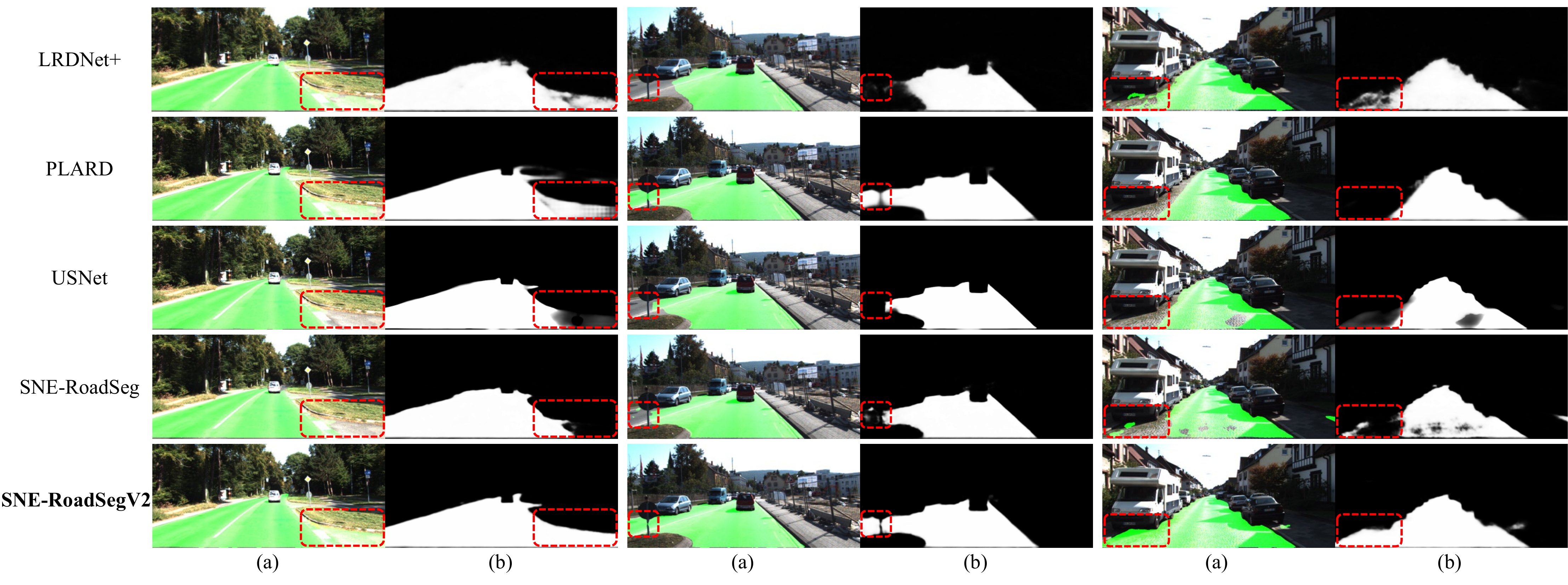}
		\caption{Qualitative comparisons of SoTA freespace detection algorithms on the KITTI Road dataset \cite{fritsch2013new}. The results of the compared algorithms are obtained using their officially published source codes and weights: (a) freespace detection results; (b) probability maps.}
		\label{fig.kitti}
	\end{figure*}
	
	\begin{table*}[!t]
		\begin{center}
			\caption{Quantitative comparison among state-of-the-art freespace detection algorithms on the KITTI Road dataset \cite{fritsch2013new}. These results are publicly available at \url{cvlibs.net/datasets/kitti/eval_road.php}. The symbol $\uparrow$ indicates the higher values correspond to the better performance, while $\downarrow$ implies the opposite. `RGB': RGB images, `Disp': disparity images, `Depth': depth images, `PC': point clouds, and `Normal': surface normal maps. }
			\label{tb.kitti}
			{
				\settablefont
				\begin{tabular}{l|c|ccccccc}
					\toprule
					Method & Input Data &{\color{purple}{\textbf{MaxF}}} (\%) $\uparrow$  & AP (\%) $\uparrow$  
					&Pre (\%) $\uparrow$  & Rec (\%) $\uparrow$   & FPR (\%) $\downarrow$ & FNR (\%) $\downarrow$  & {\color{purple}\textbf{Rank}}  $\downarrow$ \\
					\hline
					Hadamard-FCN	\cite{oeljeklaus2021integrated} &RGB 			&94.85 &91.48 &94.81 &94.89 &2.86 &5.11 &29	\\
					RBANet			\cite{sun2019reverse} 			&RGB 			&96.30 &89.72 &95.14 &97.50 &2.75 &2.50 &22	\\
					\hline
					HA-DeepLabv3+	\cite{fan2021learning} 			&RGB + Disp		&94.83 &93.24 &94.77 &94.89 &2.88 &5.11 &30	\\
					DFM-RTFNet		\cite{wang2021dynamic}  		&RGB + Disp 	&96.78 &94.05 &96.62 &96.93 &1.87 &3.07 &11	\\
					\hline
					USNet			\cite{chang2022fast} 			&RGB + Depth	&96.89 &93.25 &96.51 &97.27 &1.94 &2.73 &9	\\
					\hline
					LRDNet+			\cite{khan2022lrdnet} 			&RGB + LiDAR PC &96.95 &92.22 &96.88 &97.02 &1.72 &2.98 &8	\\
					PLB-RD          \cite{sun2022pseudo}            &RGB + LiDAR PC	&97.42 &\textbf{94.09} &97.30 &97.54 &1.49 &2.46 &4	\\ 
					PLARD			\cite{chen2019progressive}  	&RGB + LiDAR PC	&97.03 &94.03 &97.19 &96.88 &1.54 &3.12 &7	\\
					BJN				\cite{yu2021free}	 			&RGB + LiDAR PC	&94.89 &90.63 &96.14 &93.67 &2.07 &6.33 &26	\\
					LidCamNet		\cite{caltagirone2019lidar} 	&RGB + LiDAR PC	&96.03 &93.93 &96.23 &95.83 &2.07 &4.17 &18	\\
					CLCFNet			\cite{gu2021cascaded} 		&RGB + LiDAR PC	&96.38 &90.85 &96.38 &96.39 &1.99 &3.61 &15	\\
					\hline
					NIM-RTFNet		\cite{wang2020applying} 		&RGB + Normal	&96.02 &94.01 &96.43 &95.62 &1.95 &4.38 &19	\\
					SNE-RoadSeg		\cite{fan2020sne}  				&RGB + Normal	&96.75 &94.07 &96.90 &96.61 &1.70 &3.39 &12	\\
					SNE-RoadSeg+	\cite{wang2021sne}  			&RGB + Normal	&97.50 &93.98 &97.41 &97.58 &1.43 &2.42 &3	\\
					RoadFormer \cite{li2023roadformer}              &RGB + Normal	&97.50 &93.85 &97.16 &\textbf{97.84} &1.57 &\textbf{2.16} &2	\\
					\hline
					\rowcolor{gray!30}
					\textbf{SNE-RoadSegV2 (Ours)}					&RGB + Normal	&{\color{purple}\textbf{97.55}} &93.98 &\textbf{97.57} &97.53	&\textbf{1.34}	&2.47	&{\color{purple}\textbf{1}}	\\
					\bottomrule
				\end{tabular}
			}
		\end{center}
	\end{table*}

	\begin{figure*}[!t]
		\centering
		\includegraphics[width=0.99\textwidth]{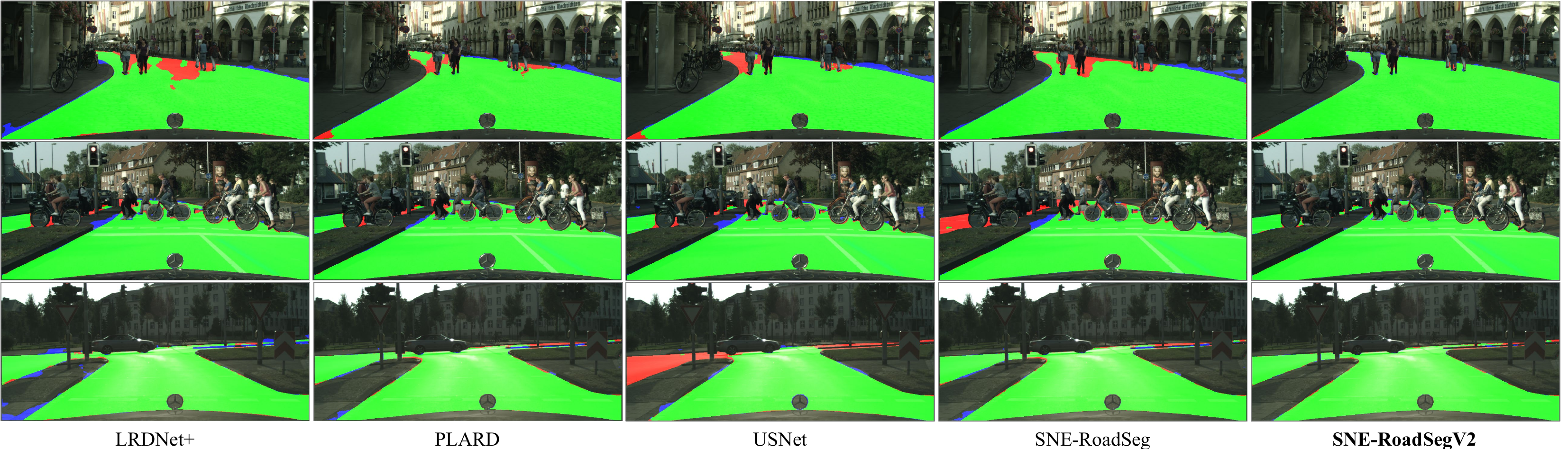}
		\caption{Qualitative comparisons of state-of-the-art freespace detection algorithms on the Cityscapes dataset \cite{cordts2016cityscapes}. The results are visualized with true-positive classifications in green, false-positive in blue, and false-negative in red.}
		\label{fig.cityscapes}
	\end{figure*}
	
	\begin{table}[!t]
		\begin{center}
			\caption{Comparison among state-of-the-art freespace detection algorithms on the Cityscapes dataset \cite{cordts2016cityscapes}.}
			\label{tb.cityscapes}
			\settablefont
			\begin{tabular}{l|ccc}
				\toprule
				Method & Fsc (\%) $\uparrow$ & IoU (\%) $\uparrow$  & Acc (\%) $\uparrow$  \\
				\hline
				NIM-RTFNet \cite{wang2020applying}				&92.02 &91.43 &96.07 \\
				RBANet \cite{sun2019reverse}					&93.81 &91.88 &96.50 \\
				USNet \cite{chang2022fast}						&94.28 &92.74 &96.71 \\
				LRDNet+ \cite{khan2022lrdnet}					&94.71 &92.82 &97.02 \\
				PLARD	\cite{chen2019progressive}				&95.28 &92.96 &97.15 \\
				SNE-RoadSeg \cite{fan2020sne}  					&96.49 &93.22 &97.68 \\
				\hline
				\rowcolor{gray!30}
				\textbf{SNE-RoadSegV2 (Ours)} 				        &\textbf{97.12} &\textbf{94.40} &\textbf{98.11} \\
				\bottomrule
			\end{tabular}
		\end{center}
	\end{table}

	\section{Experiments}
	\label{sec:experiments}
	First, we conduct extensive experiments on the KITTI road dataset \cite{geiger2012we} (medium-sized) and the Cityscapes dataset \cite{cordts2016cityscapes} (large-scale) to validate the effectiveness of our proposed encoder, decoder, and loss functions. Subsequently, we perform both qualitative and quantitative comparisons between SNE-RoadSegV2 and other SoTA freespace detection algorithms. To further demonstrate the superior performance of our network, we also provide additional experiments carried out on the vKITTI2 dataset \cite{cabon2020virtual} (large-scale yet synthetic) and the KITTI semantics dataset \cite{Alhaija2018IJCV} (real-world yet small-sized) to validate the robustness of our freespace detection framework.
	
	\subsection{Datasets and Evaluation Metrics}
	The details on the aforementioned four datasets are as follows:
	\begin{itemize}
		
		\item \textbf{KITTI Road} \cite{geiger2012we}: this dataset provides real-world RGB-D data (image resolution: 1,242$\times$375 pixels) for the evaluation of data-fusion freespace detection algorithms. Following the study presented in \cite{fan2020sne}, we split the dataset into three subsets: training (173 images), validation (58 images), and test (58 images) to conduct ablation studies and hyper-parameter selection experiments. 
		
		\item \textbf{Cityscapes} \cite{cordts2016cityscapes}: this dataset provides real-world stereo images (resolution: 2,048$\times$1,024 pixels), each manually annotated with 34 semantic classes. In our experiments, we preprocess the ground-truth annotations by categorizing them into two groups: freespace and others. As there is no depth ground truth is available, we utilize a pre-trained RAFT-Stereo \cite{lipson2021raft} to generate depth images. We adhere to the official split of training, and validation sets, with 2,975 and 500 images in each set.
		
		\item \textbf{KITTI Semantics} \cite{geiger2012we}: this dataset contains 200 real-world RGB images, each accompanied by well-annotated semantic ground truth for 19 different classes (in alignment with the Cityscapes \cite{cordts2016cityscapes} dataset). In our experiments, we pre-process the ground-truth annotations by categorizing them into two groups: freespace and others. Sparse disparity ground truth is obtained using a Velodyne HDL-64E LiDAR. We generate dense depth maps using a well-trained CreStereo \cite{li2022practical} model in our experiments. These images are randomly divided into a training set and a validation set, with a ratio of 3:1.
		
		\item \textbf{vKITTI2} \cite{cabon2020virtual}: this dataset contains virtual replicas of five sequences from the KITTI dataset. Dense ground-truth depth maps are acquired through depth rendering using a virtual engine. In our experiments, we use the first four sequences for model training, and validate the model performance on the remaining sequence.
		
	\end{itemize}
	
	\begin{figure}[!t]
		\centering
		\includegraphics[width=0.49\textwidth]{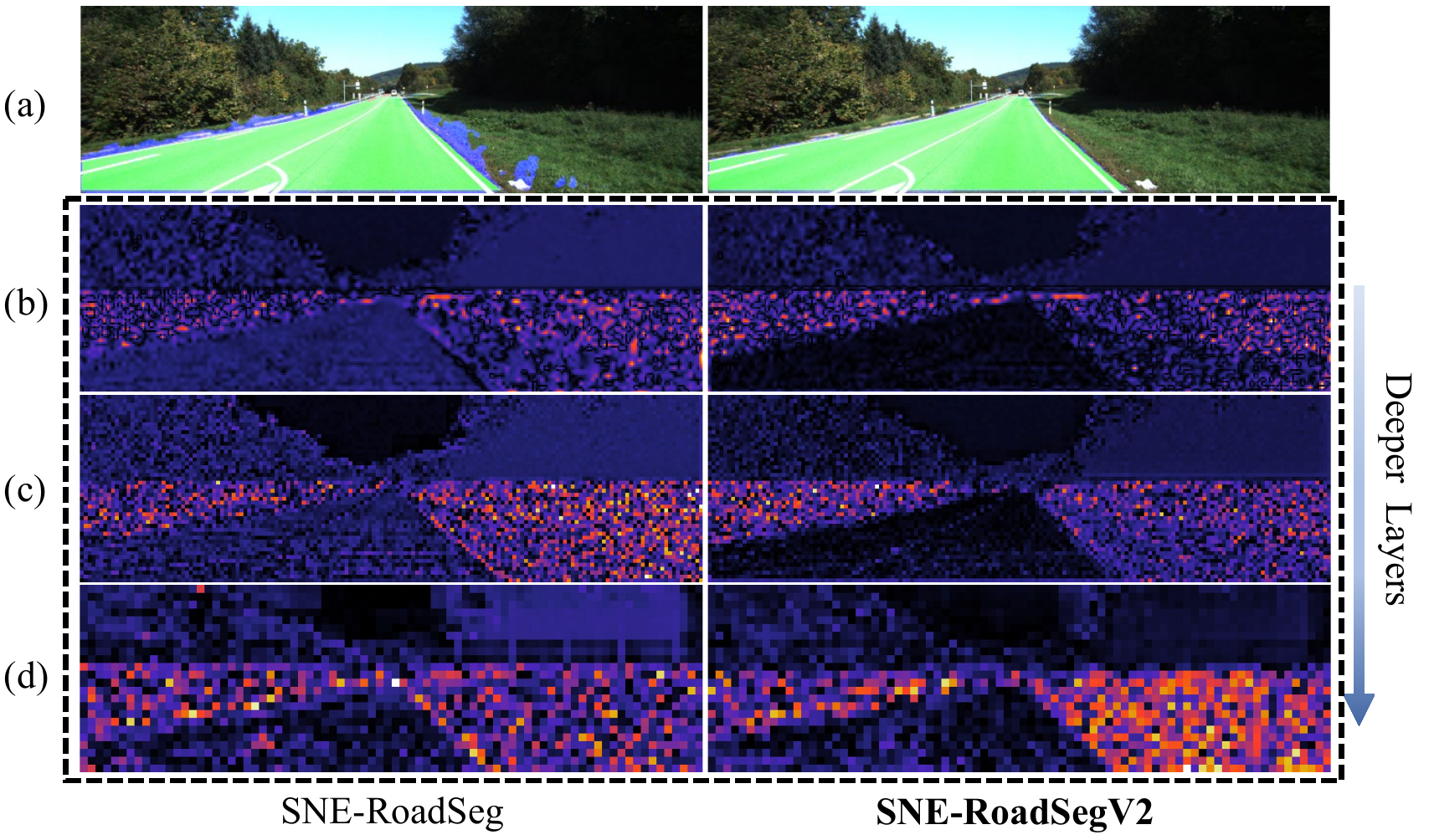}
		\caption{Visualized feature maps obtained from SNE-RoadSeg and SNE-RoadSegV2: (a) freespace detection results; (b)-(d) fused feature maps generated from shallower layers to deeper layers.}
		\label{fig.fusion_vis}
	\end{figure}
	
	Adhering to the experiments presented in \cite{fan2020sne}, we quantify the model's performance using accuracy (Acc), precision (Pre), recall (Rec), F1-score (Fsc), and intersection over union (IoU). Additionally, when submitting the results obtained from the best-performing model to the KITTI Road benchmark, we also compute maximum F1-measure (MaxF), average precision (AP), false-positive rate (FPR), and false negative rate (FNR).

	\subsection{Implementation Details} 
	\label{Sect.implementation}
	Our experiments are conducted using an Intel Core i7-12700k CPU and an NVIDIA RTX 4090 GPU. The Adam optimizer \cite{kingma2014adam} with an initial learning rate of 0.001 is used to minimize the loss function. A multi-step learning scheduler with a decay rate of 0.5 for every 20 epochs is also employed. Each model is trained for a total of 100 epochs, with early stopping mechanisms applied to the validation set to prevent over-fitting. Common data augmentation techniques, such random flipping, rotation, cropping, and brightness adjustment, are also applied to enhance the model's robustness.

	\begin{table}[!t]
		\begin{center}
			\caption{Quantitative comparison among state-of-the-art freespace detection algorithms on the KITTI Semantics dataset \cite{geiger2012we}.}
			\label{tb.kitti_semantics}
			\settablefont
			\begin{tabular}{l|ccc}
				\toprule
				Method & Fsc (\%) $\uparrow$ & IoU (\%) $\uparrow$  & Acc (\%) $\uparrow$  \\
				\hline
				NIM-RTFNet \cite{wang2020applying}				&92.59 &85.95 &96.61 \\
				OFF-Net	\cite{min2022orfd}			            &93.82 &86.79 &97.08 \\
				SNE-RoadSeg \cite{fan2020sne}  					&94.85 &88.02 &97.33 \\
				SNE-RoadSeg+ \cite{wang2021sne}  				&95.11 &89.07 &97.59 \\
				RoadFormer \cite{li2023roadformer}  			&95.36 &90.18 &97.83 \\	
				\hline
				\rowcolor{gray!30}
				\textbf{SNE-RoadSegV2 (Ours)} 				        &\textbf{96.60} &\textbf{91.75} &\textbf{98.44} \\
				\bottomrule
			\end{tabular}
		\end{center}
	\end{table}
	
	\begin{table}[!t]
		\begin{center}
			\caption{Quantitative comparison among state-of-the-art freespace detection algorithms on the vKITTI2 dataset \cite{geiger2012we}.}
			\label{tb.vkitti2}
			\settablefont
			\begin{tabular}{l|ccc}
				\toprule
				Method & Fsc (\%) $\uparrow$ & IoU (\%) $\uparrow$  & Acc (\%) $\uparrow$  \\
				\hline
				NIM-RTFNet \cite{wang2020applying}				&92.39 &85.85 &96.91 \\
				OFF-Net	\cite{min2022orfd}						&93.61 &87.53 &97.22 \\
				SNE-RoadSeg \cite{fan2020sne}  					&94.03 &88.74 &97.63 \\
				SNE-RoadSeg+ \cite{wang2021sne}  				&95.20 &90.50 &97.93 \\
				RoadFormer \cite{li2023roadformer}  			&95.39 &91.72 &98.31 \\	
				\hline
				\rowcolor{gray!30}
				\textbf{SNE-RoadSegV2 (Ours)} 				        &\textbf{96.47} &\textbf{92.33} &\textbf{98.93} \\
				\bottomrule
			\end{tabular}
		\end{center}
	\end{table}

	\subsection{Comparison with State-of-the-art Methods}
	\label{sec.comparison_with_sota}
	\begin{figure*}[!t]
		\centering
		\includegraphics[width=0.99\textwidth]{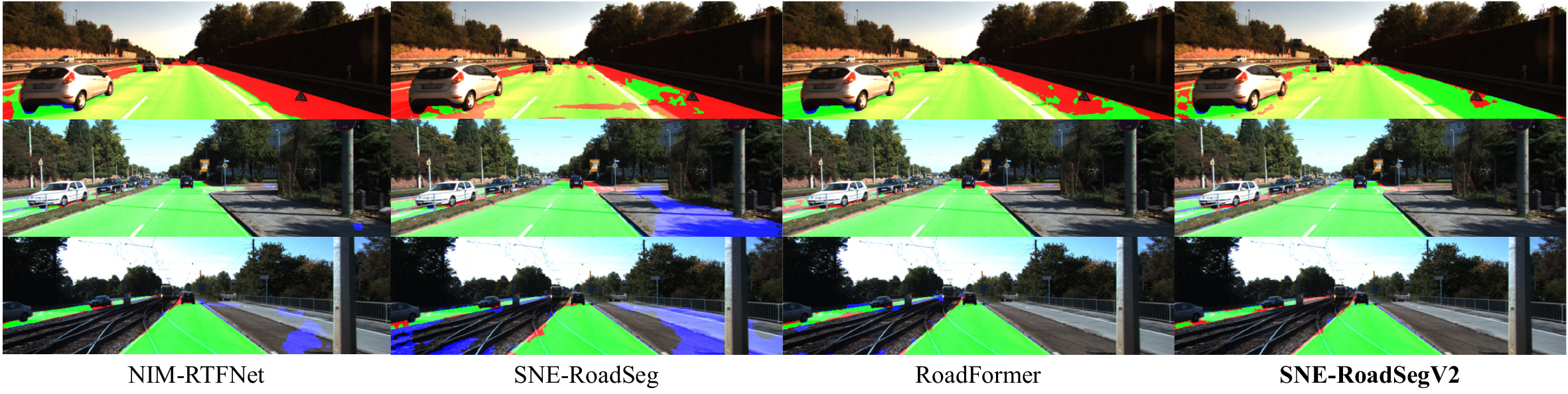}
		\caption{Qualitative comparisons of state-of-the-art freespace detection algorithms on the KITTI Semantics dataset \cite{geiger2012we}. The results are visualized with true-positive classifications in green, false-positive in blue, and false-negative in red.}
		\label{fig.kitti_semantics}
	\end{figure*}
	\begin{figure*}[!t]
		\centering
		\includegraphics[width=0.99\textwidth]{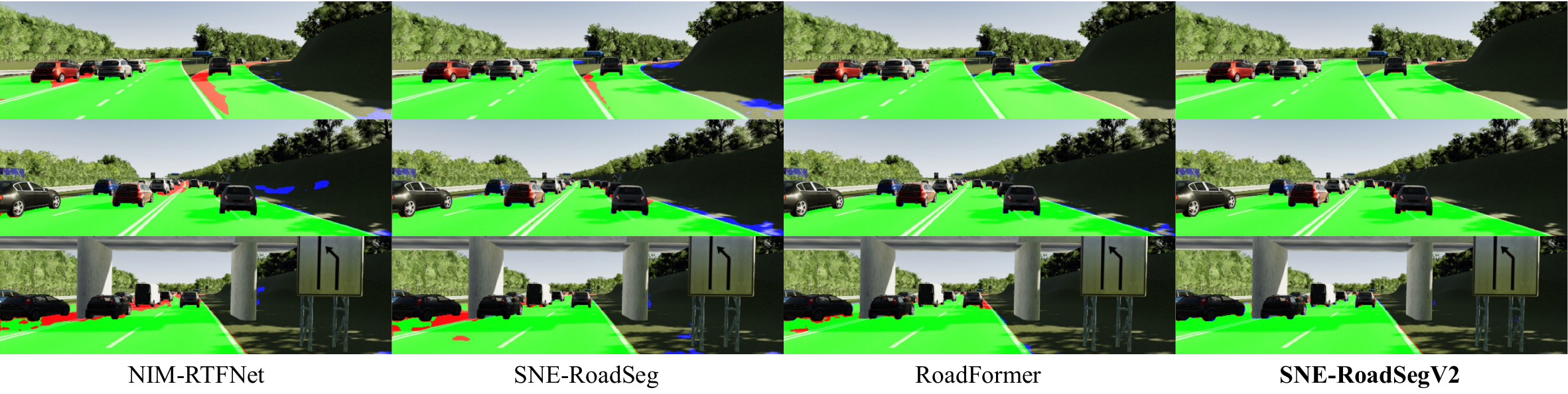}
		\caption{Qualitative comparisons of state-of-the-art freespace detection algorithms on the vKITTI2 dataset \cite{cabon2020virtual}. The results are visualized with true-positive classifications in green, false-positive in blue, and false-negative in red.}
		\label{fig.vkitti2}
	\end{figure*}

	The quantitative and qualitative experimental results on the KITTI Road dataset are presented in Table \ref{tb.kitti} and Fig. \ref{fig.kitti}, respectively, while the quantitative and qualitative experimental results on the Cityscapes dataset are given in Table \ref{tb.cityscapes} and Fig. \ref{fig.cityscapes}, respectively. These results suggest that our proposed SNE-RoadSegV2 demonstrates superior performance compared to all other SoTA freespace detection approaches, with an increase in MaxF up to 2.72\% on the KITTI dataset and an increase in IoU by 1.18-2.97\% on the Cityscapes dataset. Notably, it achieves the \textbf{1st rank} on the KITTI Road benchmark. The qualitative comparisons, with significantly improved regions highlighted by red dashed boxes, particularly near semantic-transition and depth-inconsistent regions, also validate the effectiveness of our designed feature fusion block, decoder, and loss function. The ablation studies that validate the individual efficacy of these components are discussed in the next subsection.
	
	The quantitative comparisons among SoTA freespace detection algorithms on the KITTI Semantics \cite{geiger2012we} and VKITTI2 datasets are presented in Tables \ref{tb.kitti_semantics} and \ref{tb.vkitti2}, respectively, while their qualitative comparisons on these two datasets are provided in Figs. \ref{fig.kitti_semantics} and \ref{fig.vkitti2}. These results suggest that our proposed SNE-RoadSegV2 outperforms all other SoTA approaches, with an increase in Fsc by 1.24-4.01\% on the KITTI Semantics dataset, and by 1.08-4.08\% on the VKITTI2 dataset. It is noteworthy that our SNE-RoadSegV2 demonstrates greater robustness, particularly near/on semantic transition regions or in areas with surface normals similar to that of the road surface. The achieved improvements are primarily attributed to our heterogeneous feature fusion strategy and fallibility-aware loss functions.
	
\begin{figure*}[!t]
		\centering
		\includegraphics[width=0.99\textwidth]{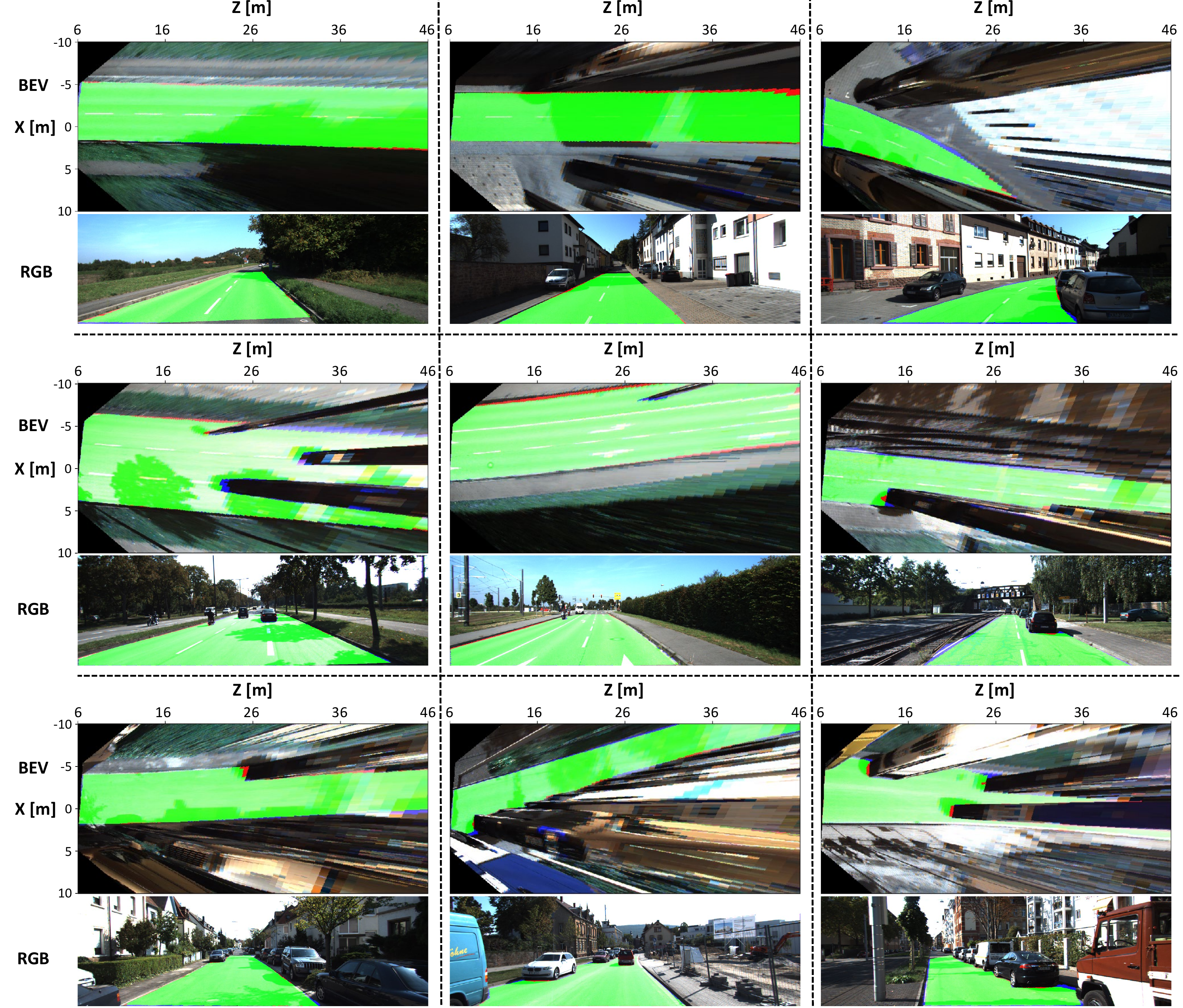}
		\caption{BEV results achieved by SNE-RoadSegV2 on the KITTI Road benchmark.}
		\label{fig.kitti_bev}
	\end{figure*}

	We also present experimental results across various road scene categories: (1) urban marked (UM), (2) urban multiple marked (UMM), and (3) urban unmarked (UU), as specified in the KITTI Road benchmark at \url{cvlibs.net/datasets/kitti/eval_road.php}. As shown in Table \ref{tb.kitti_tasks}, our proposed SNE-RoadSegV2 achieves SoTA performance in both UM and UU scenes, and demonstrates comparable performance in the UMM scene, when compared to the recent SoTA algorithm, RoadFormer \cite{li2023roadformer}. We also provide illustrative experimental results in the bird's eye view (BEV) in Fig. \ref{fig.kitti_bev}. These results clearly demonstrate the robust performance of our SNE-RoadSegV2 in detecting freespace.

	\begin{table}[!t]
		\begin{center}
			\caption{Quantitative results, reported in  MaxF (\%), across different road scene categories on the KITTI Road benchmark.}
			\label{tb.kitti_tasks}
			\settablefont
			\begin{tabular}{l|ccc}
				\toprule
				\multirow{2}*{Methods} 
				&\multicolumn{3}{c}{Road Scenes} \\
				
				\cline{2-4}
				&UM
				&UMM
				&UU \\		
				\hline
				CLCFNet \cite{gu2021cascaded}			&95.65	&97.24	&95.68	\\
				NIM-RTFNet \cite{wang2020applying}		&95.71	&96.79	&95.11	\\
				RBANet \cite{sun2019reverse}			&95.78 	&97.38	&94.91	\\
				LRDNet+ \cite{khan2022lrdnet}			&96.10	&97.98	&96.18 	\\
				SNE-RoadSeg \cite{fan2020sne}  			&96.42	&97.47	&96.03	\\
				DFM-RTFNet \cite{wang2021dynamic}       &96.46	&97.45	&96.26	\\
				USNet \cite{chang2022fast}				&96.46	&97.68	&96.11	\\
				PLB-RD \cite{sun2022pseudo}             &96.87 	&98.05	&95.63	\\
				SNE-RoadSeg+ \cite{wang2021sne}  		&96.95	&98.13	&97.04	\\
				RoadFormer \cite{li2023roadformer}      &97.02	&\textbf{98.15}	&97.02	\\
				PLARD \cite{chen2019progressive}		&97.05	&97.77	&95.95	\\
				\hline
				\rowcolor{gray!30}
				\textbf{SNE-RoadSegV2 (Ours)} 				&\textbf{97.25} 	&98.10	&\textbf{97.08}	\\
				\bottomrule
			\end{tabular}
		\end{center}
	\end{table}
	
	\begin{table}[!tb]
		\begin{center}
			\caption{Quantitative comparison of network parameters, Fsc, and IoU across various backbones.}
			\label{tb.backbone}
			{
				\settablefont
				\begin{tabular}{c|c|cc|cc}
					\toprule
					\multirow{2}*{Backbone} 
					&\multirow{2}*{Params (M)} 
					&\multicolumn{2}{c|}{KITTI Road Dataset}
					&\multicolumn{2}{c}{Cityscapes Dataset} \\
					
					\cline{3-6}
					& &
					Fsc (\%) & IoU (\%) & Fsc (\%)& IoU (\%)\\
					\hline
					Eff-B0	&\textbf{11.85}	&95.98	&92.36	&96.59	&95.57	\\
					Eff-B1	&17.22	&96.08	&92.46	&97.14	&95.05	\\
					Eff-B2	&20.29	&96.33	&92.92	&95.68	&96.99	\\
					Eff-B3	&28.00	&96.44	&93.12	&95.85	&97.03	\\
					Eff-B4	&44.78	&96.78	&93.75	&97.10	&96.46	\\
					Eff-B5	&69.46	&96.21	&92.69	&97.38	&95.06	\\
					Eff-B6	&98.83	&96.26	&92.79	&96.47	&96.05	\\
					Eff-B7	&151.80	&96.44	&93.12	&96.38	&96.49	\\
					\hline
					Res-18	&34.31	&95.87	&92.07	&96.88	&94.89	\\
					Res-34	&59.92	&95.86	&92.04	&96.84	&94.89	\\
					Res-50	&188.75	&96.27	&92.81	&96.92	&95.63	\\
					Res-101	&245.73	&96.47	&93.18	&96.76	&96.19	\\
					Res-152	&292.66	&96.92	&94.02	&96.92	&96.92	\\
					\hline
					Swin-T	&66.53	&96.34	&92.94	&96.62	&96.07	\\
					Swin-S	&109.12	&96.91	&93.26	&97.11	&97.02	\\
					\rowcolor{gray!30}
					Swin-B	&193.85	&\textbf{97.44}	&\textbf{94.13}	&\textbf{97.25}	&\textbf{97.24}	\\
					\bottomrule
				\end{tabular}
			}
		\end{center}
	\end{table}

	\begin{figure*}[!t]
		\centering
		\includegraphics[width=0.99\textwidth]{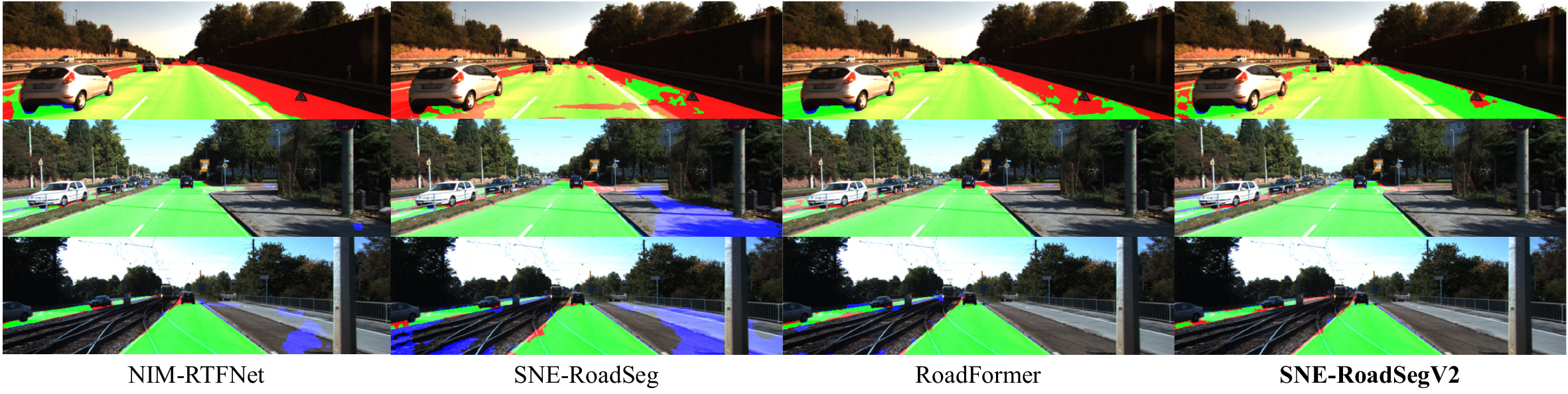}
		\caption{Qualitative comparisons of state-of-the-art freespace detection algorithms on the KITTI Semantics dataset \cite{geiger2012we}. The results are visualized with true-positive classifications in green, false-positive in blue, and false-negative in red.}
		\label{fig.kitti_semantics}
	\end{figure*}
	\begin{figure*}[!t]
		\centering
		\includegraphics[width=0.99\textwidth]{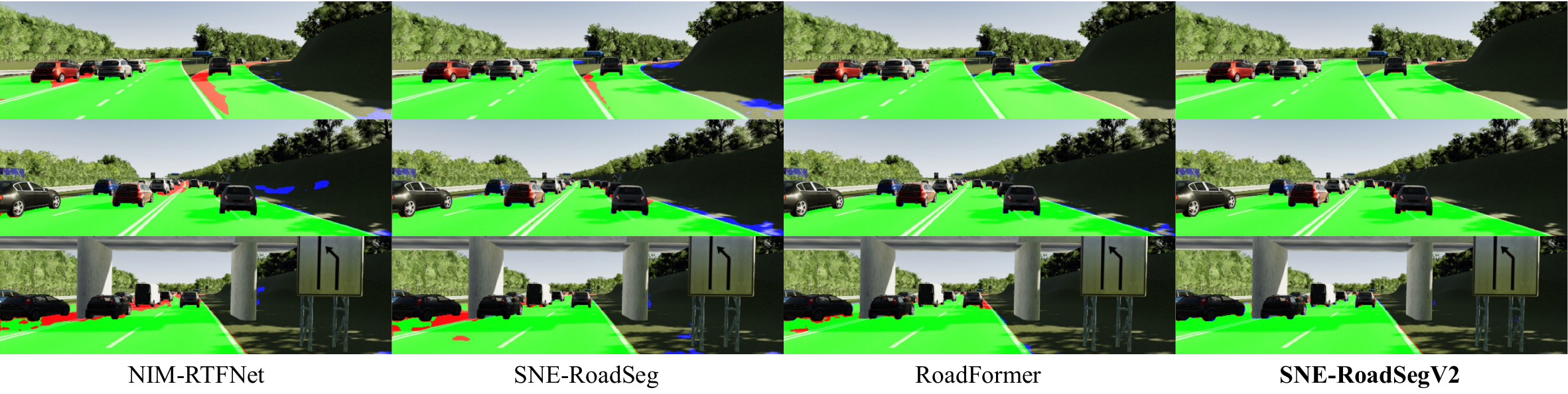}
		\caption{Qualitative comparisons of state-of-the-art freespace detection algorithms on the vKITTI2 dataset \cite{cabon2020virtual}. The results are visualized with true-positive classifications in green, false-positive in blue, and false-negative in red.}
		\label{fig.vkitti2}
	\end{figure*}

	\begin{figure}[!t]
		\centering
		\includegraphics[width=0.49\textwidth]{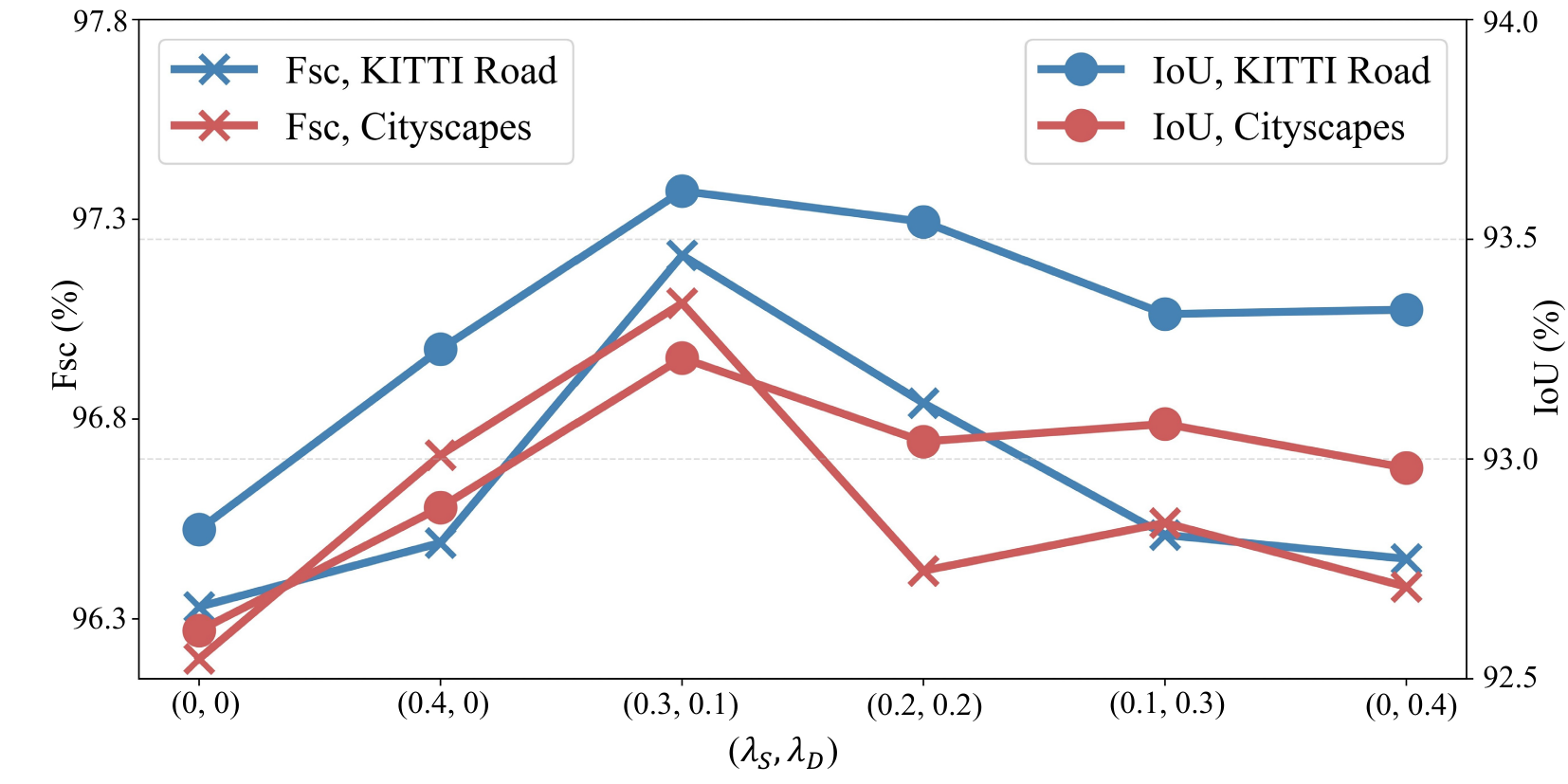}
		\caption{The selection of hyper-parameters $\lambda_S$ and $\lambda_D$ in (\ref{eq.wholeLoss}). }
		\label{fig.loss}
	\end{figure}

	\subsection{Ablation Study}
	\label{Sect.ablation}

	\begin{table*}[t!]
		\begin{center}
			\caption{Comparison of decoders in terms of both accuracy and computational complexity on the KITTI Road dataset.}
			{
				\settablefont
				\begin{tabular}{c|cccc}
					\toprule
					Decoder	& Fsc (\%) $\uparrow$ 	& IoU (\%) $\uparrow$	& Params (M) $\downarrow$ & FLOPS (G) $\downarrow$ \\
					\hline
					UNet++ \cite{zhou2019unet++} / RoadSeg \cite{fan2020sne}	&96.27	&93.90	&13.62 &78.44 \\                          	
					UNet3+ \cite{huang2020unet}	&95.64	&93.41	&14.70 &164.63 \\    
					\rowcolor{gray!30}
					\textbf{SNE-RoadSegV2 (Ours)} \cite{huang2020unet}	&\textbf{97.58}	&\textbf{94.50}	&\textbf{6.71} &\textbf{60.33} \\     
					
					\bottomrule
				\end{tabular}
			}
			\label{tb.decoder}
		\end{center}
	\end{table*}
	
	\subsubsection{Selection of Duplex Encoder Backbone}
	We evaluate the heterogeneous feature extraction performance using various backbones, including EfficientNet \cite{tan2019efficientnet}, ResNet \cite{he2016deep}, and Swin Transformer \cite{liu2021swin}. As shown in Table \ref{tb.backbone}, the backbones based on Swin Transformer outperform those based on EfficientNet and ResNet in terms of Fsc and IoU. Notably, the Swin-B backbone achieves the best overall performance among all the backbones on both the KITTI Road and Cityscapes datasets. Consequently, we select Swin-B as the network backbone for our model.

	\subsubsection{Encoder}
	We first explore the rationality of each component in HF$^2$B. As presented in Table \ref{tb.fusion_inner}, we sequentially remove each component from HF$^2$B to quantify its impact on the overall performance. It is evident that each component in HF$^2$B contributes to an improvement in the overall performance, and the network achieves its peak performance when all components (HAM, HFCD, and AWFR) are integrated into HF$^2$B, which demonstrates the effectiveness of our design. 
	
	Additionally, we compare HF$^2$B with other SoTA heterogeneous feature fusion strategies. As shown in Table \ref{tb.fusion_compare}, HF$^2$B outperforms other compared methods on both datasets, with improvements of up to 1.26\% in Fsc and 1.67\% in IoU, respectively. These compelling results can be attributed to our novel contributions, particularly the exploitation of both shared and distinct characteristics of heterogeneous features in HFCD, and the effective feature recalibration based on the affinity volume constructed in AWFR. Another qualitative comparison illustrated in Fig. \ref{fig.fusion_vis} further validates the effectiveness of HF$^2$B. Obviously, HF$^2$B produces more coherent feature representations with clearer segmentation boundaries, compared to the indiscriminate feature fusion strategy adopted in SNE-RoadSeg.

	\begin{table}[!t]
		\begin{center}
			\caption{Ablation study on the design of HF$^2$B. ``Baseline'': standard feature fusion employed in SNE-RoadSeg, ``SA'': spatial attention, ``CA'': channel attention, ``AC'': atrous convolutions.}
			\label{tb.fusion_inner}
			\settablefont
			{	
				\begin{tabular}{c|ccc|c|c|c}
					\toprule
					\multirow{2}*{\makecell{Baseline}} 
					&\multicolumn{3}{c|}{HAM}
					&\multirow{2}*{HFCD}
					&\multirow{2}*{ACFR} 
					&\multirow{2}*{Fsc (\%) $\uparrow$} \\
					
					\cline{2-4}
					&SA & CA &AC & & &\\
					\hline
					
					\hline
					\checkmark	
					&			&			&			&			&			&96.65	\\
					&\checkmark	&			&			&\checkmark	&\checkmark	&96.54	\\
					&\checkmark	&\checkmark	&			&\checkmark	&\checkmark	&97.11	\\
					\rowcolor{gray!30}
					&\textbf{\checkmark}	&\textbf{\checkmark}	&\textbf{\checkmark}	&\textbf{\checkmark}	&\textbf{\checkmark}	&\textbf{97.69}	\\
					&\checkmark	&\checkmark	&\checkmark	&			&\checkmark	&96.84	\\
					&\checkmark	&\checkmark	&\checkmark	&\checkmark	&			&97.08	\\
					\bottomrule
				\end{tabular}
			}
		\end{center}
	\end{table}

	\begin{figure}[!t]
		\centering
		\includegraphics[width=0.49\textwidth]{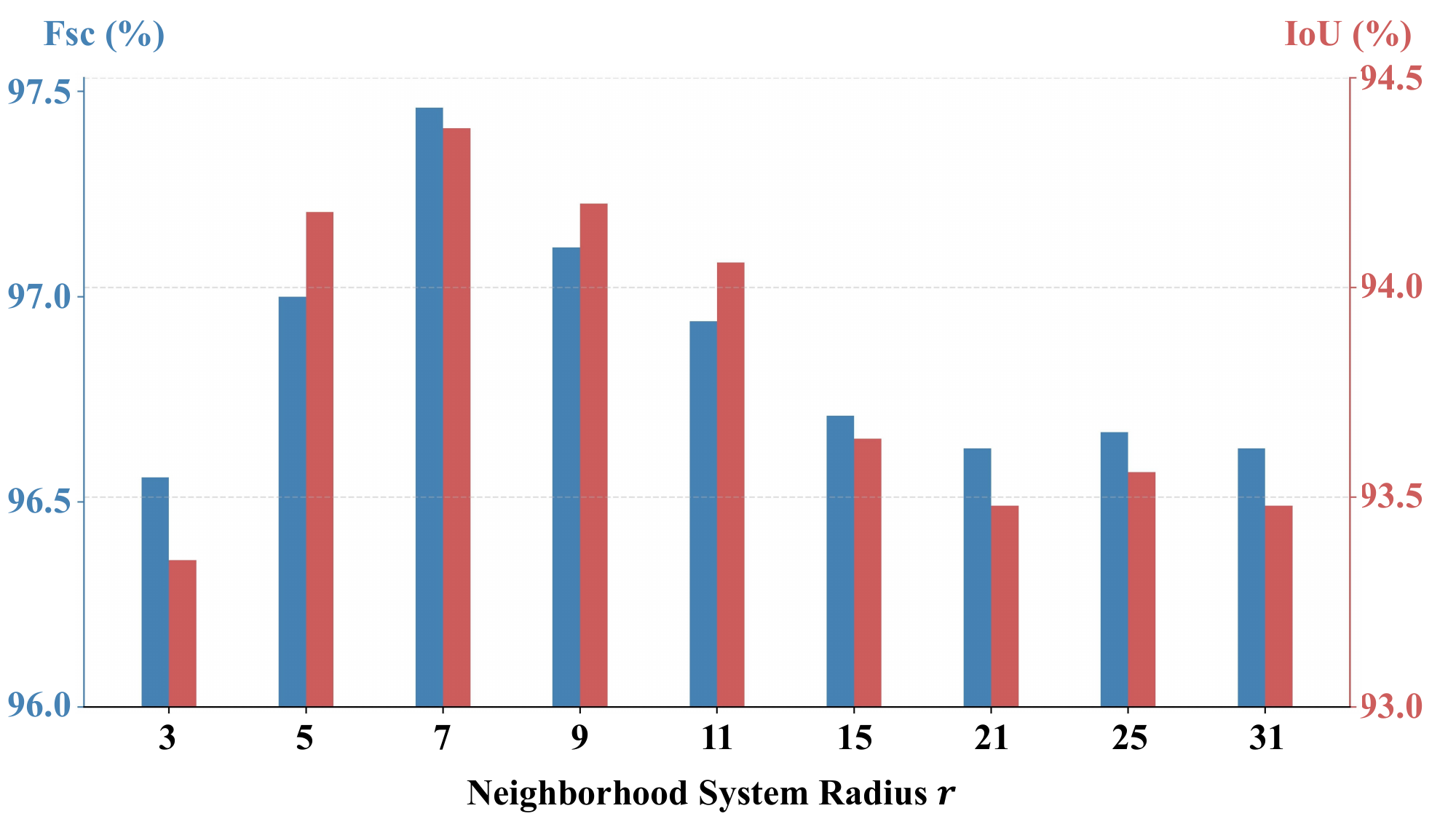}
		\caption{Analysis of network performance across various neighborhood system radii on the KITTI Road dataset.}
		\label{fig.radius}
	\end{figure}

	\begin{table}[!t]
		\begin{center}
			\caption{Comparison between our proposed HF$^2$B and other SoTA heterogeneous feature fusion strategies.}
			\label{tb.fusion_compare}
			\settablefont
			\begin{tabular}{c|cc|cc}
				\toprule
				
				\multirow{2}*{Strategy} 
				&\multicolumn{2}{c|}{KITTI Road Dataset}
				&\multicolumn{2}{c}{Cityscapes Dataset}\\
				
				\cline{2-5}
				
				&Fsc (\%) &IoU (\%)
				&Fsc (\%) &IoU (\%)\\
				\hline
				
				CFM \cite{wei2020f3net}			&95.15	&91.58	&95.45	&92.33\\
				SAGate \cite{chen2020bi}		&96.77	&93.74	&96.80	&93.81\\
				DDPM \cite{pang2020hierarchical}&96.05	&92.98	&96.60	&93.43\\
				\hline 
				\rowcolor{gray!30}
				\textbf{HF$^2$B (Ours)	}				    &\textbf{97.68}	&\textbf{94.90}	&\textbf{97.12}	&\textbf{94.40}\\
				\bottomrule
			\end{tabular}
		\end{center}
	\end{table}

	\subsubsection{Decoder}
	\label{sec.exp_decoder}
	
	Quantitative comparisons of decoder performance among SNE-RoadSegV2, RoadSeg/UNet++, and UNet3+ are presented in Table. \ref{tb.decoder}. These results demonstrate the SoTA performance of our decoder,  with improvements in Fsc and IoU by up to 1.94\% and 1.09\%, respectively, while maintaining lower computational complexity, including a reduction in learnable parameters and FLOPS by up to 54.35\% and 63.35\%, respectively. These improvements can be attributed to the use of depth-wise separable convolution and the pruning of redundant skip connections.

	\subsubsection{Loss Function}
	
	Fig. \ref{fig.loss} actually presents two experiments: (1) when $\lambda_S=\lambda_D=0$ (only conventional BCE loss is used), the overall freespace detection performance on both datasets is the worst, demonstrating the effectiveness of our proposed fallibility-aware losses; (2) different ratios between $\lambda_S$ and $\lambda_D$ demonstrate that when $\lambda_S=0.3$ and $\lambda_D=0.1$ (the sum of these two weights is empirically set to 0.4, as discussed in \cite{Huang_2019_ICCV, Li_2019_ICCV, yuan2020object}), SNE-RoadSegV2 achieves the best performance on both datasets. While further hyper-parameter tuning is possible, it is important to consider the risk of over-fitting with limited data.

	As discussed in Sec. 3.4.1, in order to maximize the effectiveness of the proposed semantic transition-aware loss $\mathcal{L}_{STA}$, we conduct experiments with a series of neighborhood system radii. As illustrated in Fig. \ref{fig.radius}, SNE-RoadSegV2 achieves the best overall performance on the KITTI Road dataset when the neighborhood radius is set to $7$ pixels.

	\section{Conclusion}
	\label{sec:conclusion}
	This paper revisited the designs of heterogeneous feature fusion strategies, decoder architectures, and loss functions from prior research, and introduced SNE-RoadSegV2, a novel, high-performing, state-of-the-art freespace detection network. Breaking down our contributions further, our work contains six technical contributions: three novel components in the encoder, one decoder architecture, and two loss functions. The effectiveness of each contribution was validated through extensive experiments. Comprehensive comparisons with other state-of-the-art algorithms unequivocally demonstrate the superiority of SNE-RoadSegV2, notably achieving the top rank on the KITTI Road benchmark. Our future work will mainly focus on extending SNE-RoadSegV2 to other general-purpose semantic segmentation tasks.

	\normalem
	\bibliographystyle{IEEEtran}
	\bibliography{ref}

\begin{thebibliography}{10}
\providecommand{\url}[1]{#1}
\csname url@samestyle\endcsname
\providecommand{\newblock}{\relax}
\providecommand{\bibinfo}[2]{#2}
\providecommand{\BIBentrySTDinterwordspacing}{\spaceskip=0pt\relax}
\providecommand{\BIBentryALTinterwordstretchfactor}{4}
\providecommand{\BIBentryALTinterwordspacing}{\spaceskip=\fontdimen2\font plus
\BIBentryALTinterwordstretchfactor\fontdimen3\font minus
  \fontdimen4\font\relax}
\providecommand{\BIBforeignlanguage}[2]{{%
\expandafter\ifx\csname l@#1\endcsname\relax
\typeout{** WARNING: IEEEtran.bst: No hyphenation pattern has been}%
\typeout{** loaded for the language `#1'. Using the pattern for}%
\typeout{** the default language instead.}%
\else
\language=\csname l@#1\endcsname
\fi
#2}}
\providecommand{\BIBdecl}{\relax}
\BIBdecl

\bibitem{fan2020sne}
R.~Fan \emph{et~al.}, ``{SNE-RoadSeg}: Incorporating surface normal information
  into semantic segmentation for accurate freespace detection,'' in
  \emph{Proceedings of the European Conference on Computer Vision
  (ECCV)}.\hskip 1em plus 0.5em minus 0.4em\relax Springer, 2020, pp. 340--356.

\bibitem{chen2019progressive}
Z.~Chen \emph{et~al.}, ``Progressive {LiDAR} adaptation for road detection,''
  \emph{IEEE/CAA Journal of Automatica Sinica}, vol.~6, no.~3, pp. 693--702,
  2019.

\bibitem{hazirbas2017fusenet}
C.~Hazirbas \emph{et~al.}, ``{FuseNet}: Incorporating depth into semantic
  segmentation via fusion-based cnn architecture,'' in \emph{Proceedings of the
  Asian Conference on Computer Vision (ACCV)}.\hskip 1em plus 0.5em minus
  0.4em\relax Springer, 2017, pp. 213--228.

\bibitem{ha2017mfnet}
Q.~Ha \emph{et~al.}, ``{MFNet}: Towards real-time semantic segmentation for
  autonomous vehicles with multi-spectral scenes,'' in \emph{2017 IEEE/RSJ
  International Conference on Intelligent Robots and Systems (IROS)}.\hskip 1em
  plus 0.5em minus 0.4em\relax IEEE, 2017, pp. 5108--5115.

\bibitem{zhou2022fanet}
X.~Zhou \emph{et~al.}, ``{FANet}: Feature aggregation network for rgbd saliency
  detection,'' \emph{Signal Processing: Image Communication}, vol. 102, p.
  116591, 2022.

\bibitem{zhou2022canet}
H.~Zhou \emph{et~al.}, ``{CANet}: Co-attention network for {RGB-D} semantic
  segmentation,'' \emph{Pattern Recognition}, vol. 124, p. 108468, 2022.

\bibitem{wang2021sne}
H.~Wang \emph{et~al.}, ``{SNE-RoadSeg+}: Rethinking depth-normal translation
  and deep supervision for freespace detection,'' in \emph{2021 IEEE/RSJ
  International Conference on Intelligent Robots and Systems (IROS)}.\hskip 1em
  plus 0.5em minus 0.4em\relax IEEE, 2021, pp. 1140--1145.

\bibitem{zou2023dual}
W.~Zou \emph{et~al.}, ``Dual geometric perception for cross-domain road
  segmentation,'' \emph{Displays}, vol.~76, p. 102332, 2023.

\bibitem{huang2020unet}
H.~Huang \emph{et~al.}, ``{UNet 3+}: A full-scale connected unet for medical
  image segmentation,'' in \emph{IEEE international Conference on Acoustics,
  Speech and Signal Processing (ICASSP)}.\hskip 1em plus 0.5em minus
  0.4em\relax IEEE, 2020, pp. 1055--1059.

\bibitem{liu2021swin}
Z.~Liu \emph{et~al.}, ``Swin {Transformer}: Hierarchical vision transformer
  using shifted windows,'' in \emph{Proceedings of the IEEE International
  Conference on Computer Vision (ICCV)}, 2021, pp. 10\,012--10\,022.

\bibitem{geiger2012we}
A.~Geiger \emph{et~al.}, ``Are we ready for autonomous driving? the {KITTI}
  vision benchmark suite,'' in \emph{2012 IEEE Conference on Computer Vision
  and Pattern Recognition (CVPR)}.\hskip 1em plus 0.5em minus 0.4em\relax IEEE,
  2012, pp. 3354--3361.

\bibitem{long2015fully}
J.~Long \emph{et~al.}, ``Fully convolutional networks for semantic
  segmentation,'' in \emph{2015 IEEE Conference on Computer Vision and Pattern
  Recognition (CVPR)}, 2015, pp. 3431--3440.

\bibitem{ronneberger2015u}
O.~Ronneberger \emph{et~al.}, ``{U-Net}: Convolutional networks for biomedical
  image segmentation,'' in \emph{Medical Image Computing and Computer-Assisted
  Intervention (MICCAI)}.\hskip 1em plus 0.5em minus 0.4em\relax Springer,
  2015, pp. 234--241.

\bibitem{badrinarayanan2017segnet}
V.~Badrinarayanan \emph{et~al.}, ``{SegNet}: A deep convolutional
  encoder-decoder architecture for image segmentation,'' \emph{IEEE
  Transactions on Pattern Analysis and Machine Intelligence}, vol.~39, no.~12,
  pp. 2481--2495, 2017.

\bibitem{chen2017deeplab}
L.-C. Chen \emph{et~al.}, ``{DeepLab}: Semantic image segmentation with deep
  convolutional nets, atrous convolution, and fully connected crfs,''
  \emph{IEEE Transactions on Pattern Analysis and Machine Intelligence},
  vol.~40, no.~4, pp. 834--848, 2017.

\bibitem{li2023roadformer}
J.~Li \emph{et~al.}, ``{RoadFormer}: Duplex transformer for rgb-normal semantic
  road scene parsing,'' \emph{CoRR}, 2023.

\bibitem{alvarez2012road}
J.~M. Alvarez \emph{et~al.}, ``Road scene segmentation from a single image,''
  in \emph{Proceedings of the European Conference on Computer Vision
  (ECCV)}.\hskip 1em plus 0.5em minus 0.4em\relax Springer, 2012, pp. 376--389.

\bibitem{xiao2016monocular}
L.~Xiao \emph{et~al.}, ``Monocular road detection using structured random
  forest,'' \emph{International Journal of Advanced Robotic Systems}, vol.~13,
  no.~3, p. 101, 2016.

\bibitem{Brust2015ConvolutionalPN}
C.-A. Brust \emph{et~al.}, ``Convolutional patch networks with spatial prior
  for road detection and urban scene understanding,'' in \emph{International
  Conference on Computer Vision Theory and Applications (VISAPP)}, 2015.

\bibitem{levi2015stixelnet}
D.~Levi \emph{et~al.}, ``{StixelNet}: A deep convolutional network for obstacle
  detection and road segmentation.'' in \emph{Proceedings of the British
  Machine Vision Conference (BMVC)}, vol.~1, no.~2, 2015, p.~4.

\bibitem{khan2022lrdnet}
A.~A. Khan \emph{et~al.}, ``{LRDNet}: Lightweight {LiDAR} aided cascaded
  feature pools for free road space detection,'' \emph{IEEE Transactions on
  Multimedia}, pp. 1--13, 2022.

\bibitem{gu2021cascaded}
S.~Gu \emph{et~al.}, ``A cascaded {LiDAR-Camera} fusion network for road
  detection,'' in \emph{2021 IEEE international conference on robotics and
  automation (ICRA)}.\hskip 1em plus 0.5em minus 0.4em\relax IEEE, 2021, pp.
  13\,308--13\,314.

\bibitem{chang2022fast}
Y.~Chang \emph{et~al.}, ``Fast road segmentation via uncertainty-aware
  symmetric network,'' in \emph{2022 International Conference on Robotics and
  Automation (ICRA)}.\hskip 1em plus 0.5em minus 0.4em\relax IEEE, 2022, pp.
  11\,124--11\,130.

\bibitem{sun2019reverse}
J.-Y. Sun \emph{et~al.}, ``Reverse and boundary attention network for road
  segmentation,'' in \emph{Proceedings of the IEEE/CVF International Conference
  on Computer Vision Workshops}, 2019, pp. 0--0.

\bibitem{wei2020f3net}
J.~Wei \emph{et~al.}, ``{F$^3$Net}: fusion, feedback and focus for salient
  object detection,'' in \emph{Proceedings of the AAAI conference on Artificial
  Intelligence (AAAI)}, vol.~34, no.~07, 2020, pp. 12\,321--12\,328.

\bibitem{chen2020bi}
X.~Chen \emph{et~al.}, ``Bi-directional cross-modality feature propagation with
  separation-and-aggregation gate for {RGB-D} semantic segmentation,'' in
  \emph{Proceedings of the European Conference on Computer Vision
  (ECCV)}.\hskip 1em plus 0.5em minus 0.4em\relax Springer, 2020, pp. 561--577.

\bibitem{pang2020hierarchical}
Y.~Pang \emph{et~al.}, ``Hierarchical dynamic filtering network for {RGB-D}
  salient object detection,'' in \emph{Proceedings of the European Conference
  on Computer Vision (ECCV)}.\hskip 1em plus 0.5em minus 0.4em\relax Springer,
  2020, pp. 235--252.

\bibitem{qiu2021semantic}
S.~Qiu \emph{et~al.}, ``Semantic segmentation for real point cloud scenes via
  bilateral augmentation and adaptive fusion,'' in \emph{2021 IEEE Conference
  on Computer Vision and Pattern Recognition (CVPR)}, 2021, pp. 1757--1767.

\bibitem{sun2020real}
L.~Sun \emph{et~al.}, ``Real-time fusion network for {RGB-D} semantic
  segmentation incorporating unexpected obstacle detection for road-driving
  images,'' \emph{IEEE Robotics and Automation Letters}, vol.~5, no.~4, pp.
  5558--5565, 2020.

\bibitem{caltagirone2019lidar}
L.~Caltagirone \emph{et~al.}, ``{LiDAR}--camera fusion for road detection using
  fully convolutional neural networks,'' \emph{Robotics and Autonomous
  Systems}, vol. 111, pp. 125--131, 2019.

\bibitem{hu2018squeeze}
J.~Hu \emph{et~al.}, ``Squeeze-and-excitation networks,'' in \emph{2018 IEEE
  Conference on Computer Vision and Pattern Recognition (CVPR)}, 2018, pp.
  7132--7141.

\bibitem{xu2015show}
K.~Xu \emph{et~al.}, ``Show, attend and tell: Neural image caption generation
  with visual attention,'' in \emph{International Conference on Machine
  Learning (ICML)}.\hskip 1em plus 0.5em minus 0.4em\relax PMLR, 2015, pp.
  2048--2057.

\bibitem{vaswani2017attention}
A.~Vaswani \emph{et~al.}, ``Attention is all you need,'' \emph{Advances in
  neural information processing systems}, vol.~30, 2017.

\bibitem{woo2018cbam}
S.~Woo \emph{et~al.}, ``Cbam: Convolutional block attention module,'' in
  \emph{Proceedings of the European Conference on Computer Vision (ECCV)},
  2018, pp. 3--19.

\bibitem{zhou2019unet++}
Z.~Zhou \emph{et~al.}, ``{UNet++}: Redesigning skip connections to exploit
  multiscale features in image segmentation,'' \emph{IEEE Transactions on
  Medical Imaging}, vol.~39, no.~6, pp. 1856--1867, 2019.

\bibitem{chollet2017xception}
F.~Chollet, ``Xception: Deep learning with depthwise separable convolutions,''
  in \emph{2017 IEEE Conference on Computer Vision and Pattern Recognition
  (CVPR)}, 2017, pp. 1251--1258.

\bibitem{fritsch2013new}
J.~Fritsch \emph{et~al.}, ``A new performance measure and evaluation benchmark
  for road detection algorithms,'' in \emph{16th International IEEE Conference
  on Intelligent Transportation Systems (ITSC)}.\hskip 1em plus 0.5em minus
  0.4em\relax IEEE, 2013, pp. 1693--1700.

\bibitem{oeljeklaus2021integrated}
M.~Oeljeklaus, \emph{An Integrated Approach for Traffic Scene Understanding
  from Monocular Cameras}.\hskip 1em plus 0.5em minus 0.4em\relax VDI Verlag,
  2021.

\bibitem{fan2021learning}
R.~Fan \emph{et~al.}, ``Learning collision-free space detection from stereo
  images: Homography matrix brings better data augmentation,'' \emph{IEEE/ASME
  Transactions on Mechatronics}, vol.~27, no.~1, pp. 225--233, 2021.

\bibitem{wang2021dynamic}
H.~Wang \emph{et~al.}, ``Dynamic fusion module evolves drivable area and road
  anomaly detection: A benchmark and algorithms,'' \emph{IEEE Transactions on
  Cybernetics}, vol.~52, no.~10, pp. 10\,750--10\,760, 2021.

\bibitem{sun2022pseudo}
L.~Sun \emph{et~al.}, ``{Pseudo}-{LiDAR}-based road detection,'' \emph{IEEE
  Transactions on Circuits and Systems for Video Technology}, vol.~32, no.~8,
  pp. 5386--5398, 2022.

\bibitem{yu2021free}
B.~Yu \emph{et~al.}, ``Free space detection using camera-{LiDAR} fusion in a
  bird's eye view plane,'' \emph{Sensors}, vol.~21, no.~22, p. 7623, 2021.

\bibitem{wang2020applying}
H.~Wang \emph{et~al.}, ``Applying surface normal information in drivable area
  and road anomaly detection for ground mobile robots,'' in \emph{2020 IEEE/RSJ
  International Conference on Intelligent Robots and Systems (IROS)}.\hskip 1em
  plus 0.5em minus 0.4em\relax IEEE, 2020, pp. 2706--2711.

\bibitem{cordts2016cityscapes}
M.~Cordts \emph{et~al.}, ``The {Cityscapes} dataset for semantic urban scene
  understanding,'' in \emph{2016 IEEE Conference on Computer Vision and Pattern
  Recognition (CVPR)}, 2016, pp. 3213--3223.

\bibitem{cabon2020virtual}
Y.~Cabon \emph{et~al.}, ``{Virtual KITTI 2},'' \emph{CoRR}, 2020.

\bibitem{Alhaija2018IJCV}
H.~Alhaija \emph{et~al.}, ``Augmented reality meets computer vision: Efficient
  data generation for urban driving scenes,'' \emph{International Journal of
  Computer Vision}, p. 961–972, 2018.

\bibitem{lipson2021raft}
L.~Lipson \emph{et~al.}, ``{RAFT-Stereo}: Multilevel recurrent field transforms
  for stereo matching,'' in \emph{2021 International Conference on 3D Vision
  (3DV)}.\hskip 1em plus 0.5em minus 0.4em\relax IEEE, 2021, pp. 218--227.

\bibitem{li2022practical}
J.~Li \emph{et~al.}, ``Practical stereo matching via cascaded recurrent network
  with adaptive correlation,'' in \emph{Proceedings of the IEEE/CVF Conference
  on Computer Vision and Pattern Recognition (CVPR)}, 2022, pp.
  16\,263--16\,272.

\bibitem{kingma2014adam}
D.~P. Kingma and J.~Ba, ``Adam: A method for stochastic optimization,''
  \emph{CoRR}, 2014.

\bibitem{tan2019efficientnet}
M.~Tan and Q.~Le, ``Efficientnet: Rethinking model scaling for convolutional
  neural networks,'' in \emph{International conference on machine
  learning}.\hskip 1em plus 0.5em minus 0.4em\relax PMLR, 2019, pp. 6105--6114.

\bibitem{he2016deep}
K.~He, X.~Zhang, S.~Ren, and J.~Sun, ``Deep residual learning for image
  recognition,'' in \emph{Proceedings of the IEEE conference on computer vision
  and pattern recognition}, 2016, pp. 770--778.

\bibitem{min2022orfd}
C.~Min \emph{et~al.}, ``{ORFD}: A dataset and benchmark for off-road freespace
  detection,'' in \emph{2022 International Conference on Robotics and
  Automation (ICRA)}.\hskip 1em plus 0.5em minus 0.4em\relax IEEE, 2022, pp.
  2532--2538.

\bibitem{Huang_2019_ICCV}
Z.~Huang \emph{et~al.}, ``{CCNet}: Criss-cross attention for semantic
  segmentation,'' in \emph{Proceedings of the IEEE/CVF International Conference
  on Computer Vision (ICCV)}, October 2019.

\bibitem{Li_2019_ICCV}
X.~Li \emph{et~al.}, ``Expectation-maximization attention networks for semantic
  segmentation,'' in \emph{Proceedings of the IEEE/CVF International Conference
  on Computer Vision (ICCV)}, October 2019.

\bibitem{yuan2020object}
Y.~Yuan \emph{et~al.}, ``Object-contextual representations for semantic
  segmentation,'' in \emph{Proceedings of the European Conference on Computer
  Vision (ECCV)}.\hskip 1em plus 0.5em minus 0.4em\relax Springer, 2020, pp.
  173--190.

\end{thebibliography}

	\clearpage

\end{document}